\documentclass[10pt,journal,compsoc,twoside]{IEEEtran}
\usepackage{hyperref}

\ifCLASSOPTIONcompsoc
  \usepackage[nocompress]{cite}
\else
  \usepackage{cite}
\fi
\usepackage{graphicx}
\usepackage{fontawesome5}
\usepackage{multirow}
\usepackage{amsmath}
\usepackage{amssymb}
\usepackage{makecell}
\usepackage{dsfont}
\usepackage[dvipsnames]{xcolor}

\colorlet{red}{black}

\definecolor{customgreen}{HTML}{95CA6D}
\definecolor{custompurple}{HTML}{B266FF}
\ifCLASSINFOpdf
\else
\fi
\begin{document}
%
\title{Integrating Affordances and Attention models for Short-Term Object Interaction Anticipation}
%
%
%
%

\author{Lorenzo~Mur-Labadia,
        Ruben~Martinez-Cantin,
        Jose~J.~Guerrero,\\
        Giovanni~Maria~Farinella,~\IEEEmembership{Senior Member,~IEEE},  and~Antonino~Furnari,~\IEEEmembership{Senior Member,~IEEE}
\IEEEcompsocitemizethanks{\IEEEcompsocthanksitem L. Mur-Labadia R. Martinez-Cantin and Jose J. Guerrero were with the Aragon Institute for Engineering Research (I3A), University of Zaragoza, Spain.\protect\\
\IEEEcompsocthanksitem G.M. Farinella and A. Furnari were with the Department of Computer Science, University of Catania, Italy.\protect\\
}
}
\IEEEtitleabstractindextext{%
\begin{abstract}
Short-Term object-interaction Anticipation (STA) consists in detecting the location of the next-active objects, the noun and verb categories of the interaction, as well as the time to contact from the observation of egocentric video. 
This ability is fundamental for wearable assistants to understand user's goals and provide timely assistance, or to enable human-robot interaction. 
In this work, we present a method to improve the performance of STA predictions. Our contributions are two-fold:
1) We propose STAformer and STAformer++, two novel attention-based architectures integrating frame-guided temporal pooling, dual image-video attention, and multiscale feature fusion to support STA predictions from an image-input video pair; 2) We introduce two novel modules to ground STA predictions on human behavior by modeling affordances. First, we integrate an environment affordance model which acts as a persistent memory of interactions that can take place in a given physical scene. We explore how to integrate environment affordances via simple late fusion and with an approach which adaptively learns how to best fuse affordances with end-to-end predictions. Second, we predict interaction hotspots from the observation of hands and object trajectories, increasing confidence in STA predictions localized around the hotspot. Our results show significant improvements on Overall Top-5 mAP, with gain up to $+23\%$ on Ego4D and $+31\%$ on a novel set of curated EPIC-Kitchens STA labels.
We released the \href{https://github.com/lmur98/AFFttention}{code, annotations, and pre-extracted affordances} on Ego4D and EPIC-Kitchens to encourage future research in this area. 
\end{abstract}

\begin{IEEEkeywords}
Short-term forecasting, Affordances, Egocentric video understanding
\end{IEEEkeywords}}

\maketitle

\markboth{IEEE Transactions on Pattern Analysis and Machine Intelligence}%
{Mur-Labadia \MakeLowercase{\textit{et al.}}: Affordances and Attention for Anticipation}

\IEEEdisplaynontitleabstractindextext

%

\ifCLASSOPTIONcompsoc
\IEEEraisesectionheading{\section{Introduction}\label{sec:introduction}}
\else
\section{Introduction}
\label{sec:intro}
\fi


\IEEEPARstart{A}{nticipating} the future is a fundamental ability for assistive egocentric devices and to support human-robot interaction. For example, a smart wearable device could alert an electrical operator before they short-circuit a switchboard, or a home robot can support the user by turning on appliances or moving objects according to their forecasted long-term goal. 
Predicting the future state of the scene from egocentric visual observations is a growing research area~\cite{plizzari2023outlook, rodin2021predicting}, 
with works tackling action anticipation~\cite{roy2024interaction, furnari2020rolling, chi2023adamsformer, nawhal2022rethinking, girdhar2021anticipative, zhong2023anticipative, zatsarynna2021multi}, locomotion prediction~\cite{lee2012discovering, park2016egocentric, bi2020can, marchetti2020multiple, kitani2012activity}, hands trajectory forecasting~\cite{liu2020forecasting, liu2022joint, bao2023uncertainty}, and next-active object detection~\cite{furnari2017next,ragusa2021meccano,jiang2021predicting,dessalene2021forecasting}.
Recently, Grauman et al.~\cite{grauman2022ego4d} defined the Short-Term Object Interaction Anticipation (STA) task as 
the simultaneous prediction of the action and object category, the object's bounding box, and the time to contact, and introduced an international challenge within the forecasting benchmark of the Ego4D dataset. Inspired by this challenge, the community proposed different approaches~\cite{chen2022internvideo, tong2022videomae, pasca2023summarize, ragusa2023stillfast, thakur2023enhancing, thakur2023guided, thakur2024leveraging}. 

\begin{figure*}[t]
\centering
\includegraphics[width=\textwidth]{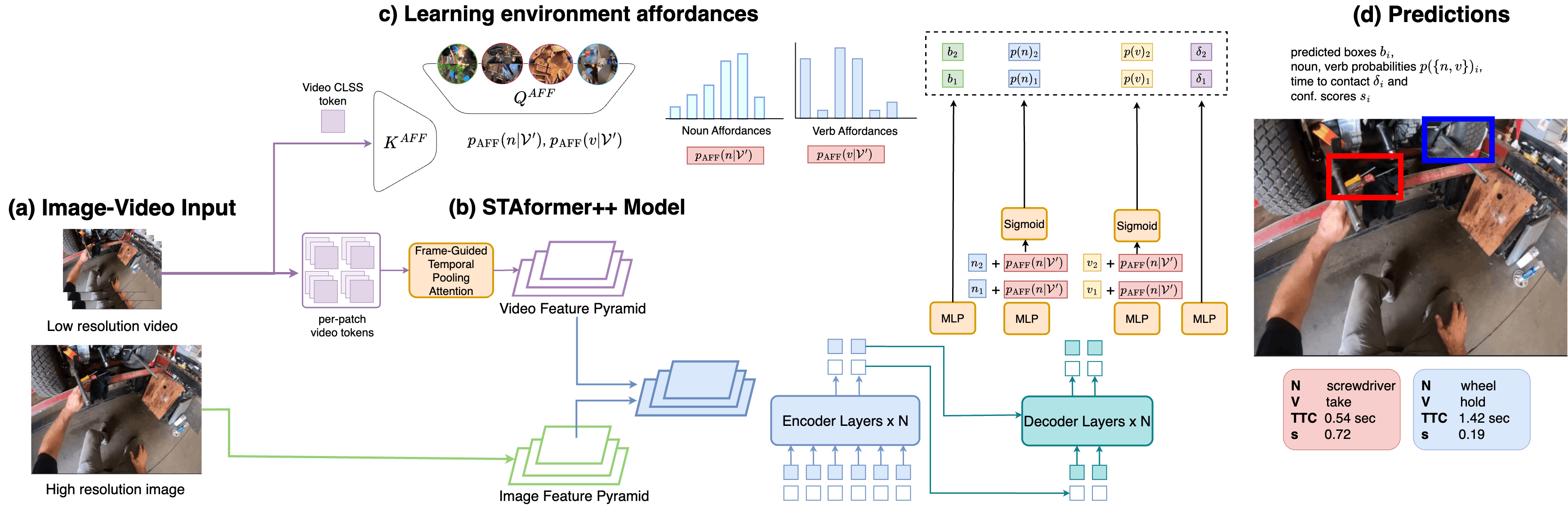}
\caption{(a) Our approach takes as input an image-video pair. (b) The input is processed by our novel STAformer++, and end-to-end short term anticipation model based on transformers which predicts object bounding boxes, the associated verb/noun probabilities, time-to-contact estimates and confidence scores. (c) \textcolor{red}{The model learns to predict environment noun and verb affordances ($p_{\text{aff}}(n \lvert \mathcal{V'})$ and $p_{\text{aff}}(v \lvert \mathcal{V'})$ in a dynamic and flexible way during training. This representation are used to refine later the predicted noun/verbs } to obtain the final predictions (e).}
\label{fig:teaser}
\end{figure*}

%
Our aim with this work is to advance research in STA with two main contributions. \textcolor{red}{An overview of our approach is presented in Figure \ref{fig:teaser}.}
First, we propose a new architectural design based on transformers to provide a principled and modern end-to-end architecture for STA which can be easily extended. 
Specifically, we introduce STAformer, which was initially proposed in our conference work~\cite{mur2025aff} and combines a transformer backbone with a Faster R-CNN detection head,
and STAformer++, which improves upon STAformer by including a novel transformer-based detection head adapted from DETR~\cite{carion2020end} for the STA task.
Differently from previous approaches~\cite{grauman2022ego4d,thakur2023guided,pasca2023summarize}, these two architectures operate on an image-video input pairs, introducing novel attention-based components for image-video fusion, such as a per-scale frame-guided temporal pooling and dual-cross attention fusion. Besides, our methods leverage the modeling capacity of state-of-the-art feature extractors such as DINOv2~\cite{oquab2023dinov2}, Swin-T~\cite{liu2021swin}, EgoVideo~\cite{pei2024egovideo} and TimeSformer~\cite{bertasius2021space}.


Second, to tackle the challenges associated with relating past visual observations to future events from video, we propose to ground predictions into human behavior by modeling environment affordances. Affordance is a psychology term coined by Gibson~\cite{gibson1977theory} as \textit{the potential actions that the environment offers to the agent}. In this work, we refer to environment affordances as the possible interactions that the agent can perform in the environment. 
As highlighted in recent studies~\cite{plizzari2023can}, human activities exhibit consistency in similar environments. Our intuition is that linking a novel video across similar environments captures a description of the feasible interactions, grounding predictions into previously observed human behavior.
We hence propose to leverage a precomputed distribution of environment affordances. By matching the input observation to our affordance database, we obtain the noun and verb affordance probabilities. During inference, these affordance distributions are used to refine the predicted verb and noun probabilities.
In a more advanced version, we integrate affordance information during training. An attention mechanism links a new video to all relevant candidates in the affordance database, enabling a more flexible approach that avoids selecting a fixed number of database members to construct the distribution.
Finally, we predict interaction hotspots~\cite{liu2022joint} to re-weigh confidence scores of STA predictions depending on the object's locations, linking predictions to spatial priors for interactions in the current frame.

The proposed approaches obtain the state-of-the-art results in the validation splits of Ego4D \cite{grauman2022ego4d} and in a novel set of curated STA annotations on the EPIC-Kitchens dataset~\cite{damen2018scaling}. 
Moreover, STAformer++ achieves competitive performance on the Ego4D Short-Term object-interaction Anticipation leaderboard.
Comparing both versions, STAformer++ achieves significant improvements over STAformer due to the combined effect of learning the affordance distribution during training and incorporating the DETR-based prediction head. Specifically, STAformer++ outperforms STAformer by +23.6$\%$ on the Ego4D v1 validation set, +10.4$\%$ on Ego4D v2, and +31.5$\%$ on EPIC-Kitchens, as measured by the official overall Top-5 mAP score.

This work is a follow-up of our previous conference paper \cite{mur2025aff}.
\textcolor{red}{The specific contributions of this extension are as follows:} 
1) \textcolor{red}{We introduce STAformer++, a novel architecture adapted to the STA task which is based on Detection Transformers.}
2) We propose to ground STA predictions in human behavior by \textcolor{red}{integrating environment affordances during training, where an attention mechanism learns to link a new video to all relevant candidates in the affordance database.} 
3) \textcolor{red}{We provide an extensive ablation of the architectural components (image and video backbones, impact of finetunning, temporal pooling, modality fusion and prediction heads), which constitute important insights about the impact of each model in a forecasting task.}
4) \textcolor{red}{Comparing the new proposed STAformer++ with STAformer, the conference version, we achieve consistent and significant relative gains across multiple datasets: +29.1 $\%$ mAP on the Ego4D-STA v1, +9.2 $\%$ mAP on the Ego4D-STA v2, +31.5 $\%$ mAP on the EPIC-Kitchens, and + 14.9 AP on the Ego4D-STA v2.}

\section{Related works}



In this section we review the key advancements in short-term object interaction anticipation, placing it within the broader context of video forecasting. We also discuss the role of affordances for anticipation, and explore the evolution of object detection architectures from convolutional to transformer-based models.

\subsection{Short-term Object Interaction Anticipation}
Furnari et al.~\cite{furnari2017next} initially introduced the concept of Next-Active Objects (NAO), proposing to detect future interacted objects by analyzing their trajectories as observed from the first-person point of view. 
Differently from action anticipation~\cite{damen2018scaling}, the NAO detection task is designed to provide grounded predictions in the form of bounding boxes, which can be particularly informative for wearable AI assistants or embodied robotic agents.
Unlike traditional object detection~\cite{girshick2015fast}, NAO prediction requires the ability to model the dynamics of the scene and anticipate the user's intention. 
Jiang et al.~\cite{jiang2021predicting} developed a method to predict the next-active object location in the form of a Gaussian heatmap from a single RGB image, combining visual attention with probabilistic maps of hand locations. Ego-OMG~\cite{dessalene2021forecasting} segments the NAO and predicts the interaction time using a contact anticipation map that captures scene dynamics. 
While previous works considered different task formulations and evaluation approaches, Grauman et al.~\cite{grauman2022ego4d} formalized NAO prediction by introducing the STA task and an associated challenge on the EGO4D dataset~\cite{grauman2022ego4d}. The initial baseline is composed of a Faster R-CNN branch to detect objects~\cite{girshick2015fast} and a SlowFast 3D CNN~\cite{feichtenhofer2019slowfast} for video processing.
Subsequent research introduced architectural enhancements and alternative approaches. Chen et al.~\cite{chen2022internvideo} employed pre-computed object detections using a DETR model and substituted SlowFast with a VideoMAE pre-trained ViT~\cite{tong2022videomae}. Pasca et al.~\cite{pasca2023summarize} proposed TransFusion, which employs a language encoder for action context summary, performing multi-modal fusion with visual features.
While previous works leveraged pre-extracted object detections for 2D image understanding, Ragusa et al.~\cite{ragusa2023stillfast} introduced StillFast, an end-to-end framework unifying the processing of 2D images and video in a combined backbone.
Thakur et al.~\cite{thakur2023enhancing} proposed GANO, an end-to-end model based on a transformer architecture including a novel guided attention mechanism. 
Guided attention was integrated within a StillFast architecture in~\cite{thakur2023guided}, achieving state-of-the-art results. Thakur et al.~\cite{thakur2024leveraging} introduced NAOGAT, a multi-modal transformer that attends detected objects and includes a motion decoder to track object trajectories. 
Recently, a video-language foundation model denominated EgoVideo  \cite{pei2024egovideo} achieved the state-of-the-art in the STA task. The authors selected 7M ego video-text clips from multiple datasets and trained the model with standard video-text contrastive learning. The video encoder was then finetunned to the STA task using the StillFast \cite{ragusa2023stillfast} prediction head.
Compared with previous works, we propose a novel architecture that fuses the image-video pair with attention-based components and that integrates affordances for refining the predictions.

\subsection{Affordances for Anticipation}
The computational perception of affordances has been investigated in different forms. 
A line of works predicts affordance labels of object parts, requiring strong supervision in the form of manually annotated masks~\cite{mur2023bayesian, do2018affordancenet, myers2015affordance, nguyen2017object}. However, these methods are not ``grounded'' in human behavior as the annotator declares interaction regions outside of any interaction context~\cite{nagarajan2019grounded}.

Other works considered the problem of grounding affordance regions in images by leveraging videos depicting human-object interactions in a weakly supervised way, where only the action label is used as supervision without spatial annotations~\cite{nagarajan2019grounded,luo2023learning,goyal2022human,li2023locate}. Nagarajan et al.~\cite{nagarajan2019grounded} introduced the concept of ``interaction hotspots'' as the potential spatial regions where the action can occur.
Mur-Labadia et al.~\cite{mur2023multi} create a 3D multi-label mapping of affordances extracted from egocentric video.
Another line of work infers interaction hotspots from video by forecasting future hand movements to select candidate regions for future interactions~\cite{jiang2021predicting, liu2022joint, liu2020forecasting, goyal2022human}.
Few works studied scene affordances to predict a list of likely actions that can be performed in a given scene~\cite{rhinehart2016learning,nagarajan2020ego}. In particular, Nagarajan et al.~\cite{nagarajan2020ego} proposed Ego-Topo, a procedure to decompose a set of egocentric videos into a topological map encoding scene affordances.
Despite the interest in affordances, only a few works investigated how to exploit them for future predictions. Montesano et al.~\cite{montesano2008learning} predicted affordance effects for human-robot interaction. Koppula et al.~\cite{koppula2015anticipating} used object affordances to anticipate human behavior in the form of motion trajectories of objects and humans.
Nagarajan et al.~\cite{nagarajan2020ego} showed how scene affordances learned from egocentric video can improve long-term action anticipation.
Liu et al.~\cite{liu2020forecasting} tackled action anticipation by jointly predicting egocentric hand motion, interaction hotspots, and future actions.
Liu et al.~\cite{liu2022joint} highlighted how interaction hotspots predicted by forecasting hand motion can support action anticipation.
In this work we integrate affordances in an unified architecture for the short therm anticipation task by the first time. In accordance to literature, we show that affordances are beneficial for performance due to its generalization capabilities. Moreover, we study how to use then during training time.

\subsection{Object Detection Architectures}
Object detectors based on convolutional networks are categorized as either two-stage or one-stage models, relying on hand-crafted anchors or reference points for object localization, respectively.
Two-stage detectors \cite{ren2016faster, he2017mask, chen2019hybrid} involve a Region Proposal Network (RPN) that generates boxes candidates that are subsequently refined. Faster-RCNN \cite{ren2016faster} applies a Region of Interest (RoI) alignment and a set of linear layers for accurate prediction of bounding boxes and semantic class for object detection. 
One-stage approaches, such as YOLO \cite{redmon2016you} directly predict offset from predefined anchors without the proposal stage, notably reducing the inference time. 
However, convolutional models still require manual components like Non-Maximun Suppression (NMS) to eliminate redundant boxes and rely heavily on anchor generation methods, affecting overall performance.

These limitations were solved by the arrival of the DEtection TRansformer (DETR) \cite{carion2020end}, an end-to-end transformer-based architecture for object detection.
DETR introduces the concept of ``object queries'', a fixed number of learned embeddings decoded to predict objects in an image, eliminating the need for hand-crafted components. During training, these queries interact with the image encoded features through cross-attention in the transformer decoder. Since each object query ultimately corresponds to a potential detected object, DETR applies simple linear layers to predict the bounding boxes and the class labels for each object.
However, the instability in the the Hungarian algorithm for matching the targets with the object queries, the lack of inductive biases like anchor boxes and the global attention mechanism make very slow the convergence of DETR.
Deformable DETR \cite{zhu2020deformable} focuses on selecting a set of sampling points and applying a deformable attention that attends to a small set of points around the sampled point, improving both the efficiency and accuracy of the model. 
DAB-DETR \cite{liu2022dab} formulates the positional part of the decoder queries as dynamic 4D anchor box coordinates $(x,y,h,w)$, which provides a reference query point $(x,y)$ and a reference anchor size $(w,h)$ that simplifies the refinement process.
DN-DETR \cite{li2022dn} introduces a De-Noising (DN) training strategy that accelerates the DETR convergence by solving the instability of the bipartite matching. It feeds noisy ground truth samples into the decoder and trains to recover the original, uncorrupted data with an additional denoising loss.
DINO-DETR \cite{zhang2022dino} combines DAB-DETR and DN-DETR with the deformable attention for its computational efficiency, including a contrastive denoising training, a mixed query selection and a novel ``look forward twice'' scheme, achieving significant improvements both in accuracy and convergence. In this work, we benchmark both convolutional and transformer based heads with DINO-v2 \cite{oquab2023dinov2} and Swin Transformer (Swin-T)\cite{liu2021swin} features. We also highlight the importance of the video encoder for modeling the action dynamics, and the importance of the intermediate components for fusing both modalities in order to obtain the better predictions.

\begin{figure*}[t]
\centering
\includegraphics[width=\textwidth]{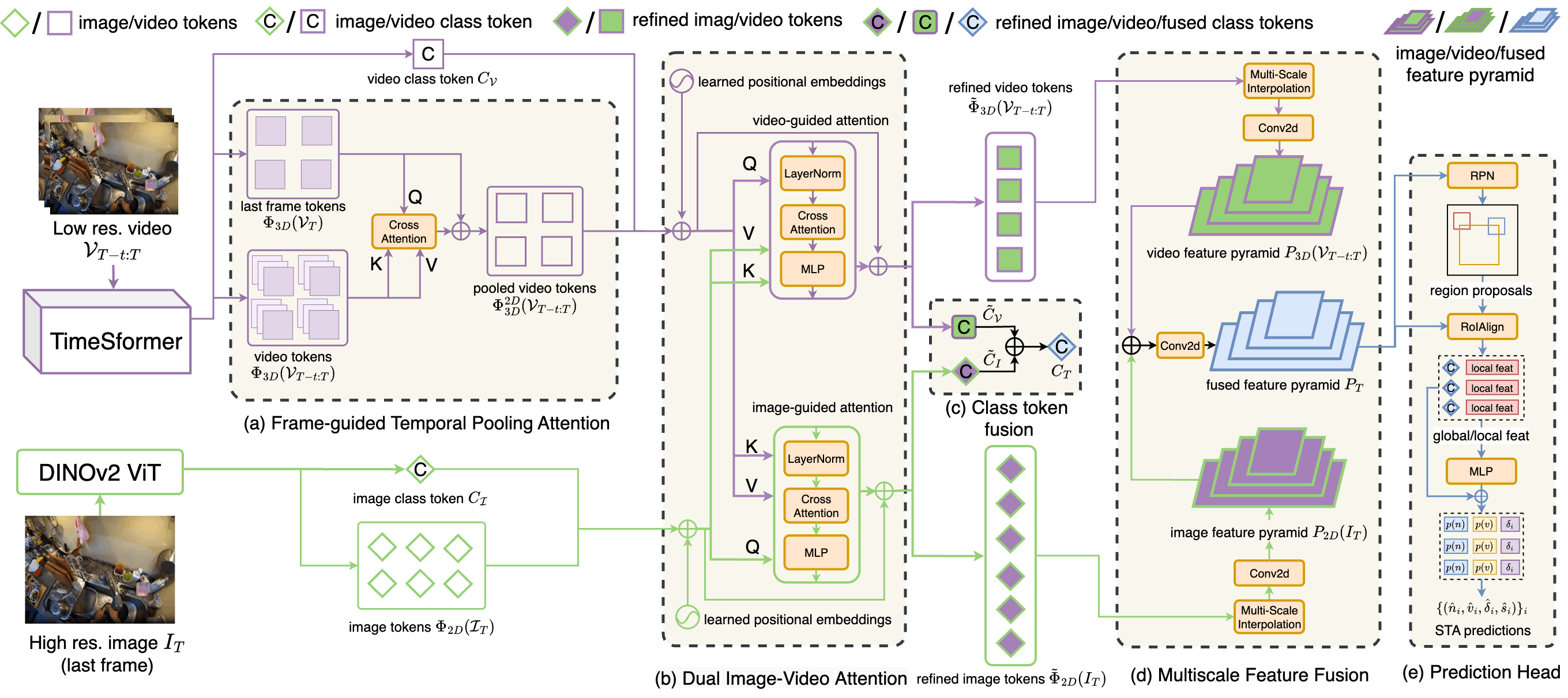}
\caption{\textbf{STAformer architecture.} DINO-v2 and TimeSformer extract 2D and 3D features form the image-video input. (a) Frame-guided temporal pooling attention spatially aligns video to image features. (b) Dual image-video attention enriches 2D features with temporal dynamics and 3D features with fine-grained image details. Image and video representations are joined to obtain a global class token (c) and a feature pyramid (d), from which we obtain the STA predictions (e).} 
\label{fig:encoder}
\end{figure*}

\section{STAformer, a Transformer-based Architecture for Short-Term Anticipation}

STAformer is a novel architecture that leverages pre-trained transformer models for image and video feature extraction~\cite{oquab2023dinov2,pramanick2023egovlpv2} and introduces novel attention-based components for image-video representation fusion. 

\subsection{Problem Formulation}

As defined in \cite{grauman2022ego4d}, the goal of Short-Term object interaction Anticipation (STA) is to detect the next-active object from the observation of the image frame at time $T$, $I_T \in \mathbb{R}^{h_s \times w_s \times c}$, and sequence of frames $\mathcal{V}_{T - t:T} \in \mathbb{R}^{t \times h_f \times w_f \times c}$ taken $t$ time-steps before $T$. The model's predictions are a set of detections, defined as a tuple $(b_m,n_m,v_m,\delta_m, s_m)$, denoting future interacted objects in the last observed frame $I_T$. Each bounding box $b_m$ is associated with an object category label $n_m$ (noun), a verb label indicating the interaction mode $v_m$, a time-to-contact $\delta_m$ indicating that the interaction will take place at time $T+\delta_m$, and a confidence score $s_m$.



\vspace{1mm}
\noindent
\subsection{Feature Extraction}
Following previous work~\cite{grauman2022ego4d,ragusa2023stillfast}, we process a high resolution image $I_T \in \mathbb{R}^{h_s \times w_s \times 3}$ sampled from the input video $\mathcal{V}_{:T}$ at time $T$ 
and a sequence of low-resolution frames $\mathcal{V}_{T - t:T} \in \mathbb{R}^{t \times h_f \times w_f \times 3}$ taken $t$ time-steps before $T$.
First, we extract high-resolution 2D features from $I_T$ with a DINOv2 model\cite{oquab2023dinov2}, obtaining a set of 2D image tokens $\Phi_{2D}(I_T)$ and a class token $C_I$ offering a global representation of the image. The high-level semantics and dense localization of DINOv2 features makes them very suitable for the object detection task. We also extract spatio-temporal 3D features from $\mathcal{V}_{T - t: T}$ using a TimeSformer model~\cite{bertasius2021space}, obtaining a set of video tokens $\Phi_{3D}(\mathcal{V}_T)$ and a class token $C_\mathcal{V}$ that captures a global representation of the input clip.

\begin{figure*}[t]
\centering
\includegraphics[width=\textwidth]{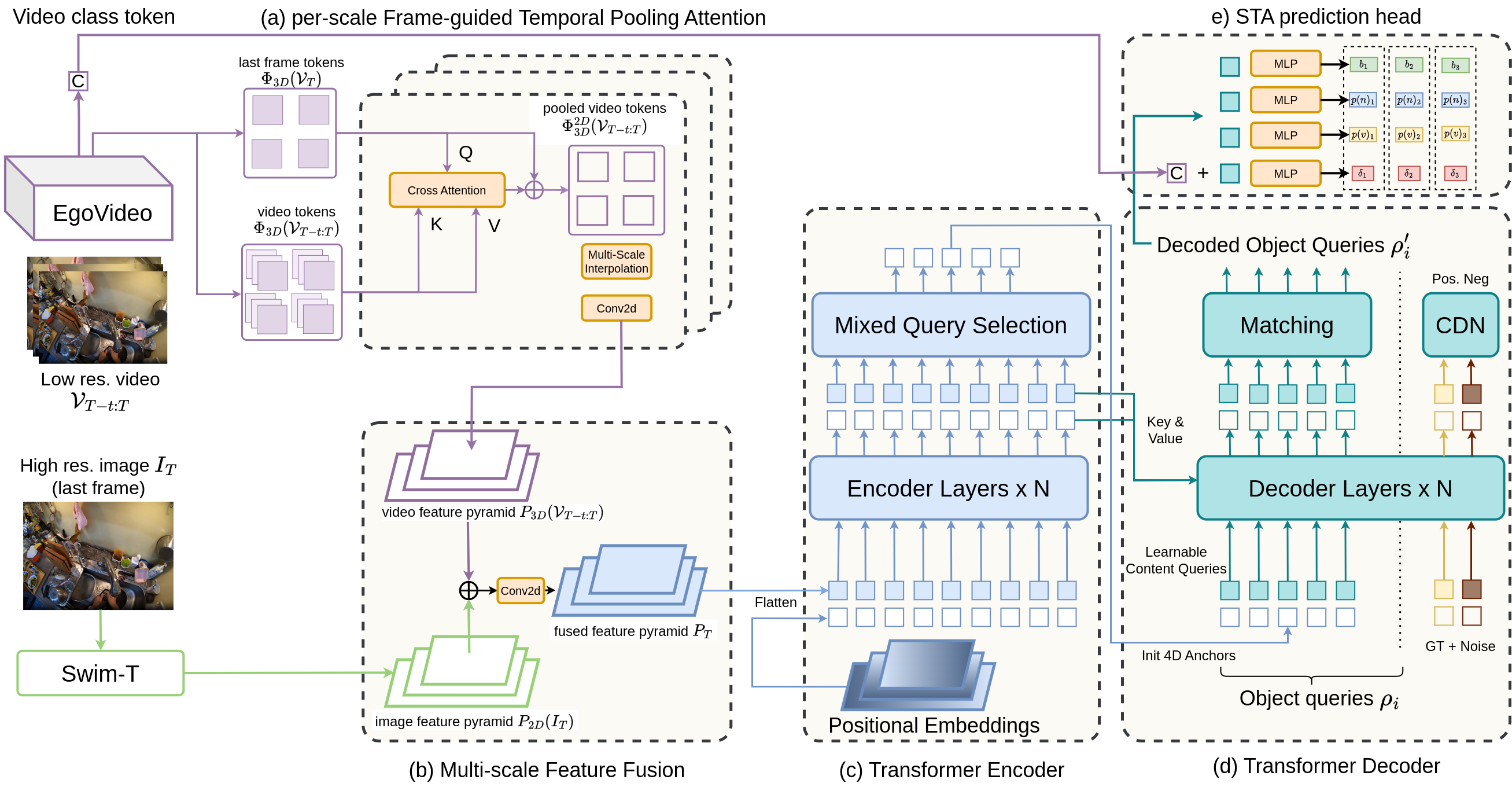}
\caption{\textbf{STAformer++ architecture.} The Swin-T backbone extracts hierarchical multi-scale 2D feature maps from the high-resolution image, while the EgoVideo backbone extracts spatio-temporal 3D features. a) We compute per-scale Frame-guided temporal pooling, and then resize the pooled video tokens to the respective image map. b) The two feature maps are summed to obtain the fused feature pyramid $P_T$. c) The DETR Encoder enhances the features and applies the Mixed Query Selection to initialize the positional part of the object queries $\rho_m$, while the content parts are kept as learnable parameters. d) The DETR Decoder incorporates the refined image-video features to the object queries. We accelerate the convergence using a Contrastive DeNoising (CDN) part with positive and negative samples as proposed in \cite{li2022dn}. e) The STA prediction head applies independent MLP layers to obtain the final predictions $(\hat b_m, \hat n_m, \hat v_m, \hat \delta_m, \hat s_m)$.} 
\label{fig:encoder_detr}
\end{figure*}

\noindent
\subsection{Frame-guided Temporal Pooling Attention (Figure~\ref{fig:encoder}(a))}
While the overall video tokens provide a spatio-temporal representation of the input video, STA predictions need to be aligned to the spatial location of the last video frame. The frame-guided temporal pooling attention maps video tokens to the spatial reference system of the last video frame, compressing the 3D representation obtained by the TimeSformer to a 2D one. 
The 3D video tokens $\Phi_{3D}(\mathcal{V}_{T-t:T})$ are mapped to 2D pooled video tokens denoted as $\Phi_{3D}^{2D}(\mathcal{V}_{T-t:T})$ adopting a residual cross-attention mechanism. 
Specifically, we compute query vectors from last-frame video tokens $\Phi_{3D}(\mathcal{V}_T)$ with a linear projection $W_{Q}$, while key and value vectors are computed from the overall video tokens $\Phi_{3D}(\mathcal{V}_{T-t:T})$ using the $W_K$ and $W_V$ linear projection layers. 
We obtain pooled video tokens with a residual multi-head attention ($A$) layer as follows:

\begin{equation}
\begin{aligned}
\Phi_{3D}^{2D}(\mathcal{V}_{T-t:T}) &= \Phi_{3D}(\mathcal{V}_T) + A(Q_{TP}, K_{TP}, V_{TP}) \\
Q_{TP} &= \Phi_{3D}(\mathcal{V}_{T}) W_{Q} \\
K_{TP} &= \Phi_{3D}(\mathcal{V}_{T - t:T})W_{K} \\
V_{TP} &= \Phi_{3D}(\mathcal{V}_{T - t:T})W_{V}
\end{aligned}
\end{equation}

\noindent
Used as queries, last-frame tokens guide an adaptive temporal pooling that summarizes the spatio-temporal video feature map computed and maps it to the 2D reference space of the last observed frame. The residual connection facilitates learning and lets the attention mechanism focus on enriching last-frame tokens with video tokens.

\noindent
\subsection{Dual Image-Video Attention fusion (Figure~\ref{fig:encoder}(b))}
Image tokens $\Phi_{2D}(I_T)$ and pooled video tokens $\Phi_{3D}^{2D}(\mathcal{V}_{T-t:T})$ are spatially aligned, but carry different information, with image tokens encoding fine-grained visual features and video tokens encoding scene dynamics.
This module adopts a residual dual cross-attention that aims to enrich image tokens with scene dynamics information coming from video tokens through image-guided cross-attention and, vice versa, video tokens with fine-grained visual information coming from image tokens through video-guided cross-attention. 
Prior to forwarding image and video tokens to the multi-head cross-attention modules, these are summed with learnable positional embeddings to capture insightful spatial relationships and normalized through a Layer Norm. The residual image-guided cross-attention is as follows:

\begin{equation}
\begin{aligned}
[\tilde{\Phi}_{2D}(I_T), \tilde{C}_I] & = [\Phi_{2D}(I_T),C_I] + A(Q_{CA}, W_{CA}, V_{CA}) \\
Q_{CA} &= [\Phi_{2D}(I_T),C_I]W_{Q} \\
K_{CA} &= [\Phi_{3D}^{2D}(\mathcal{V}_{T-t:T}),C_\mathcal{V}]W_{K} \\
V_{CA} &= [\Phi_{2D}^{3D}(\mathcal{V}_{T-t:T}),C_\mathcal{V}]W_{V}
\end{aligned}
\end{equation}



\noindent
where $[\cdot,\cdot]$ denotes concatenation along batch dimension, and $W_Q$, $W_K$, and $W_V$ are linear projection layers. After the multi-head attention layer, the refined image representation $[\tilde{\Phi}_{2D}(I_T), \tilde{C}_I]$ is passed through a residual MLP. The video-guided cross-attention works in a similar way to compute refined video tokens $\tilde{\Phi}_{3D}(\mathcal{V}_{T-t:T})$ and video class tokens $\tilde{C}_\mathcal{V}$, but queries are computed from video tokens while keys and values are computed from image tokens.

\vspace{1mm}
\noindent
\subsection{Feature Fusion and Fast-RCNN based STA prediction head (Figure~\ref{fig:encoder}(c)-(e)):}
Refined image and video class tokens are summed to obtain the overall class token $C_T = \tilde{C}_{I} + \tilde{C}_{\mathcal{V}}$, a global representation of the input image-video pair (Figure~\ref{fig:encoder}(c)). 
\textcolor{red}{Following \cite{dosovitskiy2020image}, we use the CLS token as the global representation of the scene instead of applying global average pooling.}
Refined image tokens $\tilde{\Phi}_{2D}(I_T)$ are mapped to a multi-scale feature pyramid~\cite{lin2017feature} $P_{2D}(I_T)$ by rescaling $\tilde{\Phi}_{2D}(I_T)$ to multiple resolutions using bilinear interpolation, followed by a $3 \times 3$ convolution to compensate for interpolation artifacts.
Refined video tokens $\tilde{\Phi}_{3D}(\mathcal{V}_{T-t:T})$ are mapped to a feature pyramid $P_{3D}(\mathcal{V}_{T-t:T})$ in the same way.
The two feature pyramids are summed and passed through a 2D $3 \times 3$ convolution to obtain the fused feature pyramid $P_T$ (Figure~\ref{fig:encoder}(d)). 
We adopt the prediction head proposed in Stillfast~\cite{ragusa2023stillfast} to obtain the final predictions $(\hat b_m, \hat n_m, \hat v_m, \hat \delta_m, \hat s_m)$, which modifies the Faster-RCNN~\cite{girshick2015fast} head integrating components specialized for STA prediction. In short, $P_T$ is passed to a Region Proposal Network (RPN), which computes object proposals. Such proposals are then used to extract local features from $P_T$ with RoI Align~\cite{he2017mask}, mapping bounding boxes to appropriate layers of the pyramid following~\cite{lin2017feature}. Each extracted local feature vector is concatenated with the fused class token $C_T$ and passed through an MLP with a residual connection. Linear layers are used to compute noun probabilities $p(n)_m$, verb probabilities $p(v)_m$ and time-to-contact (ttc) predictions. Note that while~\cite{ragusa2023stillfast} uses global average pooling to obtain a global representation of the scene, we naturally use the class token $C_T$ learned from the input image-video pair.




\section{STAformer++: End-to-End Short-Term Anticipation with Transformers}

While STAformer delivers state of the art performance, it still makes use of components based on convolutional object detectors, notably in the detection head, which may limit its performance. 
We investigate whether the inclusion of a transformer-based detection head can further improve performance and propose STAformer++, a redesign of the original STAformer architecture.
Specifically, we substitute the Fast-RCNN STA head by a prediction head based on DETR. We also replace the DINOv2 \cite{oquab2023dinov2} image features by Swin-T \cite{liu2021swin}, a multi-scale image transformer. We subsequently compute per-scale the frame-guided temporal pooling to pool more robust temporal features to the object size. The TimeSformer \cite{pramanick2023egovlpv2} video feature extraction is substituted by EgoVideo \cite{pei2024egovideo}, the state-of-the-art in multiple Ego4D challenges.
The following subsections detail the different STAformer++ architecture components, shown in Figure \ref{fig:encoder_detr}.


\noindent
\subsection{Feature extraction}
We process the high resolution image $I_T$
with Swin-T \cite{liu2021swin} to extract hierarchical multi-scale feature maps $P_{2D}(I_T)$. Swin-T alternates window multi-head self-attention with shifted window partitioning attention, which introduces cross-window connections. Since its computational cost grows linearly, it is a great candidate for dense vision tasks with high input image resolution. 
We use EgoVideo \cite{pei2024egovideo} for extracting the video tokens $\Phi_{3D}(\mathcal{V}_{T-t:T})$ and the video class token $C_\mathcal{V}$ from the low-resolution video  $\mathcal{V}_{T - t: T}$.

\noindent
\subsection{Per-Scale Frame-guided Temporal Pooling Attention and Feature Fusion (Figure~\ref{fig:encoder_detr}(a-b))}
The per-scale frame-guided temporal pooling attention maps video tokens to the spatial reference system of the last video frame by implicitly considering the scale of the feature map. Specifically, we repeat the frame-guided temporal pooling attention for each scale of the image features. As detailed in Section 3.2, we adopt a residual cross-attention mechanism between the video tokens of the last frame $\Phi_{3D}(\mathcal{V}_{T})$ and the full stack of 3D video features $\Phi_{3D}(\mathcal{V}_{T - t: T})$. The pooled video tokens are followed by a bilinear interpolation and a 2D Convolution to obtain the video feature pyramid $P_{3D}(\mathcal{V}_{T - t: T})$.
Then, we sum the two multi-scale feature maps to obtains the fused feature pyramid $P_T$.

\noindent
\subsection{Detection Transformer (Figure~\ref{fig:encoder_detr}(c-d))}


We flatten the fused feature map $P_T$ to obtain a sequence of tokens which are then forwarded to the DETR Encoder. We sum a fixed positional encoding to incorporate the spatial relationships of the patches. The DETR Encoder consists of standard multi-head self-attention layers followed by feed-forward networks. The self-attention mechanism of the encoder aggregates context from the entire image.

Then, the DETR Decoder processes the object queries $\rho$. We follow the mixed query selection strategy proposed by Liu et al. \cite{liu2022dab} to dynamically initialize the positional part of the object queries, while its content part remains static to accelerate the convergence. Specifically, the positional part are 4D anchor boxes composed by the reference query points  $(x,y)$ and anchor sizes $(w,h)$, obtained after a query selection of the top-K encoder features. 
We apply the deformable attention \cite{zhu2020deformable} to layer-by-layer integrate the comprehensive context from the image-video into the object queries.
We further accelerate the convergence by feeding noise-altered ground-truth labels and boxes into the DETR Decoder \cite{liu2022dab}, which trains the model for accurate ground-truth reconstruction. 
Moreover, as proposed by Zhang et al. \cite{zhang2022dino}, we adopt the Contrastive DeNoising (CDN) to discard irrelevant anchors and the ``look forward twice'' for more efficient training.

\subsection{DETR based STA prediction head (Figure~\ref{fig:encoder_detr}(e))}

From the processed object queries $\rho'$, we obtain the final predictions  $(\hat b_{\textcolor{red}{m}}, \hat n_{\textcolor{red}{m}}, \hat v_{\textcolor{red}{m}}, \hat \delta_{\textcolor{red}{m}}, \hat s_{\textcolor{red}{m}})$.
Bounding-box coordinates are computed with a 3-layer Multi-Layer Perceptron (MLP), predicting the normalized center, height, and width relative to the input image.
The noun $p(n)_m$ and verb probabilities $p(v)_m$ are predicted with two independent 3-layer MLP followed by a Sigmoid function, and considering an additional special class label $\emptyset$, which indicates that no object is detected.
The score $s_m$ of the joined prediction is the multiplication of the respective noun and verb probabilities. 
Finally, we concatenate the class token of the video model $C_\mathcal{V}$ with the decoded object queries $\rho'$  for explicitly incorporating the action dynamics. We regress the time-to-contact $ttc_i$ with a final 3-layer MLP.

\begin{figure*}[t]
\centering
\includegraphics[width=0.99\textwidth]{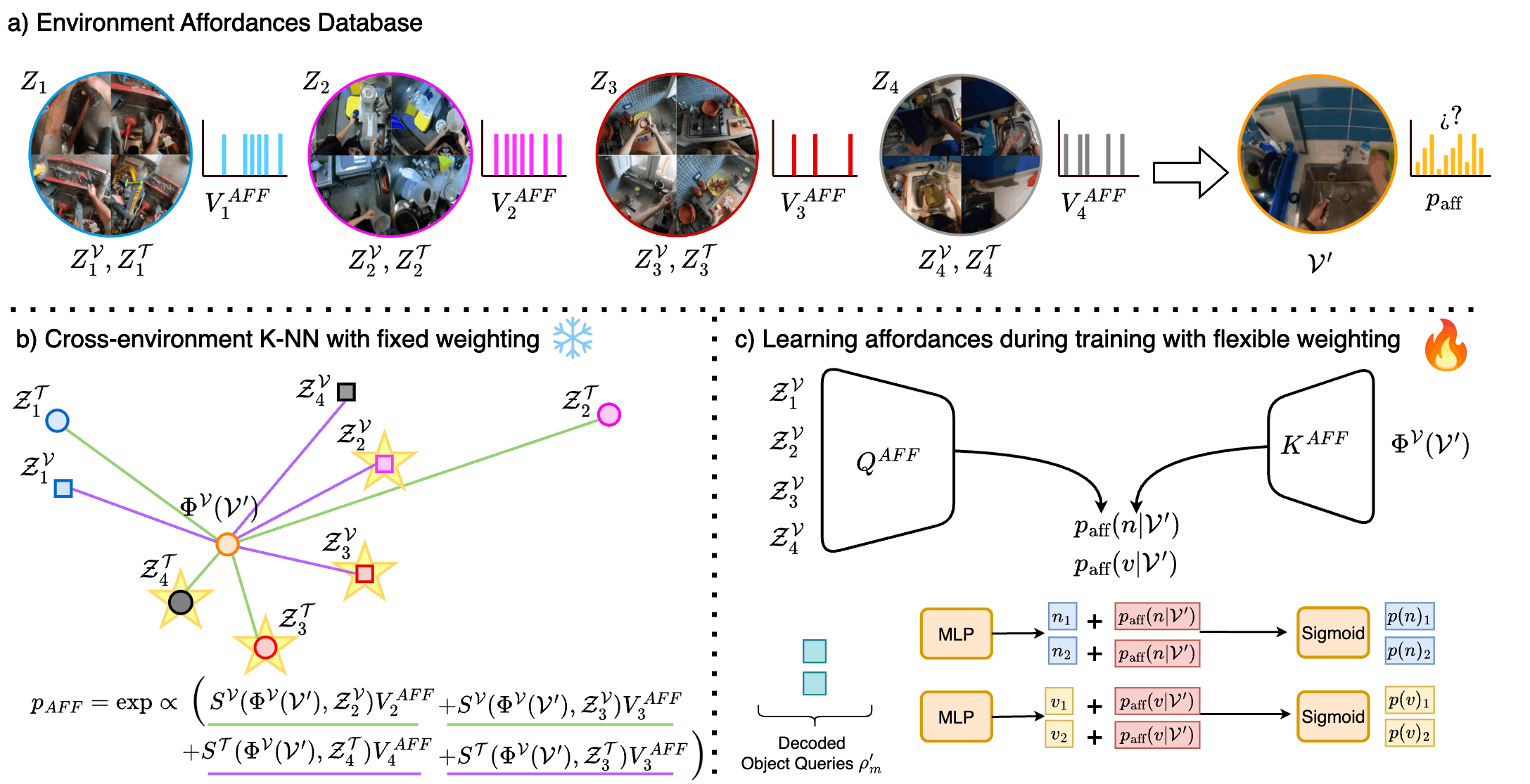}
\caption{\textbf{Environment affordances in forecasting.} 
a) We build an affordance database by linking training videos according to their visual similarity, obtaining activity-centric zones with affordances values $V_{\textcolor{red}{j}}^{AFF}$ and respective video $\mathcal{Z}_j^\mathcal{V}$, text $\mathcal{Z}_j^\mathcal{T}$ descriptors. 
b) Our first approach matches the input encoded video $\Phi^\mathcal{V}(\mathcal{V}')$ to the affordance database by selecting the K nearest neighbors in terms of the cosine similarity with the visual
$\mathcal{Z}^\mathcal{V}$ and text $\mathcal{Z}^\mathcal{V}$ zone descriptors. The affordance probability $p_{AFF}$ is obtained by weighting the counts of nouns present in the top-2K nearest zones  (\scalebox{2}{$\star$}) according to the respective similarity $\mathcal{S}$. This will be late-fused with the predictions made by the end-to-end model. Example for K=2. 
c) In our second methodology, an attention mechanism ($Q^{AFF}, K^{AFF}$) learns to associate a novel video $\mathcal{V'}$ with all the potential zone candidates $Z_j$ in the affordance database. This dynamically obtains the noun $\mathcal{N}_{AFF}$ and verb $\mathcal{A}_{AFF}$ affordance distributions, which are summed to the DETR predicted nouns $n_m$ and verb $v_m$ logits during model training. The final binary class probabilities  $p(n)_m$, $p(v)_m$ are obtained after a Sigmoid layer.}
\label{fig:aff_scheme}
\end{figure*}

\section{Environment affordances for human behavior grounding}

While end-to-end STA architectures predict the next interaction directly from input video, in this section we show that it is beneficial to ground the predictions on past observed human behavior.
Environment affordances \cite{nagarajan2020ego} refer to all potential interactions that can be performed in a given physical zone. By learning a map of the environment affordances, using egocentric videos of human activities, we are able to guide the next interaction prediction using the similarities and correlations among human activities and scenarios. 
We describe how to build an affordance semantic map (Figure \ref{fig:aff_scheme}-a)) by grouping and connecting similar training videos. Then, we present two methods for grounding predictions on environment affordances. Our first solution (Figure \ref{fig:aff_scheme}-b)) pre-computes a fixed affordance distribution of the current scene based on the affordance distributions of similar scenes or videos, where we use the cosine similarity. This affordance distribution is used during inference to refine noun and verb probabilities. Our second proposed strategy (Figure \ref{fig:aff_scheme}-c)) learns the affordance distribution during training using a flexible attention mechanism.



\subsection{Building a persistent memory of affordances}






We start extracting activity-centric zones from the training set following~\cite{nagarajan2020ego} in order to build an affordance map as Figure \ref{fig:aff_scheme}-a) shows. Each affordance zone is a group of image-video pairs with high visual similarity in a certain environment.
We create positive and negative frame pairs labels by counting homography estimation inliers, evaluating temporal coherence, and computing visual similarity. We consider two frames similar if they are less than 15 frames apart or if they share 10 inlier homography key-points. We extract SuperPoint keypoint descriptors \cite{detone2018superpoint} and use RANSAC for the homography estimation. We measure the visual similarity from pre-trained ResNet-152 features \cite{he2016deep} to select dissimilar frames.
Based on the positive and negative pairs, we train a Siamese network $\mathbb{L}$, composed by a Resnet-18 \cite{he2016deep} followed by a 5 layer multi-layer perceptron (MLP), on these pairs and used to predict the probability $\mathbb{L}(I, I')$ that two frames $I$ and $I'$ belong to the same zone. We then process all frames in a video sequence with $\mathbb{L}$ to group video frames according to their visual similarity in different zones.

Each zone $Z_{\textcolor{red}{j}}$ represents an activity-centric region composed of the group of visually similar images $I_i^Z$, their corresponding videos $\mathcal{V}_i^Z$, the associated narrations $\mathcal{T}_i^Z$, sets of nouns $\mathcal{N}_i^Z$ and action verbs $\mathcal{A}_i^Z$ appearing at least once in the STA annotations of all images $I_i^Z$, \textcolor{red}{where $i$ indexes videos within the zone}. We define the affordance distribution in each zone as a unnormalized distribution $V_{\textcolor{red}{j}}^{\mathcal{N}_{AFF}} = \mathds{1}_{n\in Z_i}$, $V_{\textcolor{red}{j}}^{\mathcal{A}_{AFF}} = \mathds{1}_{a\in Z_i}$ that considers if the noun $\mathcal{N}$, verb $\mathcal{A}$ appears in the zone ${\textcolor{red}{j}}$.
Since each zone captures the different interactions that the user performed in that specific environment, this database represents a sort of \textit{persistent memory} on how humans behave in each space.
We obtain a visual descriptor $\mathcal{Z}_{\textcolor{red}{j}}^\mathcal{V}$ and a text descriptor $\mathcal{Z}_{\textcolor{red}{j}}^\mathcal{T}$ for each zone $Z_{\textcolor{red}{j}}$ using video language pre-trained models \cite{pramanick2023egovlpv2, pei2024egovideo} to extract the zone video $\mathrm{\Psi}^\mathcal{V}(\mathcal{V}_i^Z)$ and text $\mathrm{\Psi}^\mathcal{T}(\mathcal{\mathcal{T}}_i^Z)$ descriptors as follows, \textcolor{red}{where $j$ indexes the different zones and $i$ the number of videos within each zone.}:
\begin{equation}
    \mathcal{Z}_{\textcolor{red}{j}}^\mathcal{V} = \sum_{i = 1}^{|Z|} \mathrm{\Psi}_\mathcal{V}(\mathcal{V}_i^Z) / |Z|, \quad \mathcal{Z}_{\textcolor{red}{j}}^\mathcal{T} = \sum_{i = 1}^{|Z|} \mathrm{\Psi}_\mathcal{T}(\mathcal{\mathcal{T}}_i^Z) / |Z|
\end{equation}

\noindent
\subsection{Fixed pre-inferred environment affordances}
\label{subsec:fixed_aff}

At inference time, we predict the nouns and verbs affordance distribution by matching a novel video $\mathcal{V}'$ to zones related to functionally similar environments in the affordance database. Since we can only extract a visual descriptor from the novel video, $\mathrm{\Psi}^\mathcal{V}(\mathcal{V}')$, we compute the visual cosine similarity $\mathcal{S}^\mathcal{V}(\mathrm{\Psi}^\mathcal{V}(\mathcal{V}'), \mathcal{Z}^\mathcal{V})$ and the video-text cross cosine similarity $\mathcal{S}^\mathcal{T}(\mathrm{\Psi}^\mathcal{V}(\mathcal{V}'), \mathcal{Z}^\mathcal{T})$ between the clip and each zone $Z$ in the database. 
Beyond retrieving visually similar zones, the video-text cross cosine similarity relates different locations with similar functionality that affords the same interaction (i.e, painting a wall in India or painting a canvas with watercolor in Spain both afford to dip the brush in the paint).
As illustrated in Figure~\ref{fig:aff_scheme} a), we employ the K-Nearest Neighbor algorithm to identify the most similar zones to the given input $\mathcal{V}'$. We define the top-K visual zones $\mathcal{K}^{\mathcal{V}}$, where $S_k^\mathcal{V}$ is a shorthand notation for $S^\mathcal{V}_k(\Psi(\mathcal{V}'), \textcolor{red}{\mathcal{Z}}_k^\mathcal{V})$, and the top-K narrative zones $\mathcal{K}^{\mathcal{T}}$. 

\begin{equation}
\begin{split}
    \mathcal{K}^{\mathcal{V}} =\{(\textcolor{red}{\mathcal{Z}}^\mathcal{V}_1, S^\mathcal{V}_1), ..., (\textcolor{red}{\mathcal{Z}}^\mathcal{V}_K,S^\mathcal{V}_K) \}, \\
    \mathcal{K}^{\mathcal{T}}=\{(\textcolor{red}{\mathcal{Z}}^{\mathcal{T}}_1, S^\mathcal{T}_1), ..., (\textcolor{red}{\mathcal{Z}}^{\mathcal{T}}_K,S^\mathcal{T}_K) \}
\end{split}
\end{equation}

Combining both sets, \textcolor{red}{$\mathcal{K} = \mathcal{K}^\mathcal{V} \cup \mathcal{K}^\mathcal{T} = \{ (\mathcal{Z}_k, S_k) \}_{k=1}^{2K}$} yields a total of  $2K$ zones and their respective similarity scores, which we assume to share affordances with $\mathcal{V}'$. We then define the probability of each noun $p_{\text{aff}}\left(n|\mathcal{V}'\right)$ as an exponential distribution by weighting the noun and verb appearance in each neighboring zone according to the respective similarity \textcolor{red}{$S_k$}, \textcolor{black}{where the exponential function reflects a standard softmax formulation, enabling probabilistic interpretation of affinity scores $S_k \cdot V_k^{\mathcal{N}_{AFF}}$ as logits}:

\begin{equation}
\textcolor{red}{p_{\text{aff}}\left(n|\mathcal{V}'\right) \propto \exp (
\sum_{(Z_k, S_k) \in \mathcal{K}} S_k \cdot V_k^{\mathcal{N}_{AFF}} )    }
\end{equation}

\begin{equation}
\textcolor{red}{p_{\text{aff}}\left(v|\mathcal{V}'\right) \propto \exp (
\sum_{(Z_k, S_k) \in \mathcal{K}} S_k \cdot V_k^{\mathcal{A}_{AFF}} ) }   
\end{equation}



Based on the environment affordances, we can predict probability distributions over \textit{possible} nouns $p_{\text{aff}}\left(n|\mathcal{V}\right)$ or verbs $p_{\text{aff}}\left(v|\mathcal{V}'\right)$ given past interactions in functionally similar zones. Differently, the STA model will predict probability distributions of given nouns and verbs being the next interactions  $p_{\text{sta}}\left(n|\mathcal{V}', I'\right)$ and $p_{\text{sta}}\left(v|\mathcal{V}', I'\right)$ directly from the input image-video pair, without explicitly considering the set of possible actions. 
We assume independence between the two predictions\footnote{In practice, we build the two models with different architectures and training objectives to make the dependence weak.} and perform data fusion by computing the unnormalized joint likelihoods:

\begin{equation}
\begin{aligned}
    p_{\text{fus}}(n|I', \mathcal{V}') &\propto p_{\text{aff}}\left(n|\mathcal{V}'\right) \cdot p_{\text{sta}}\left(n|\mathcal{V}', I'\right) \\
    p_{\text{fus}}(v|I', \mathcal{V}') &\propto p_{\text{aff}}\left(v|\mathcal{V}'\right) \cdot p_{\text{sta}}\left(v|\mathcal{V}', I'\right)
\end{aligned}
\end{equation}

\begin{figure*}[t]
\centering
\includegraphics[width=\textwidth]{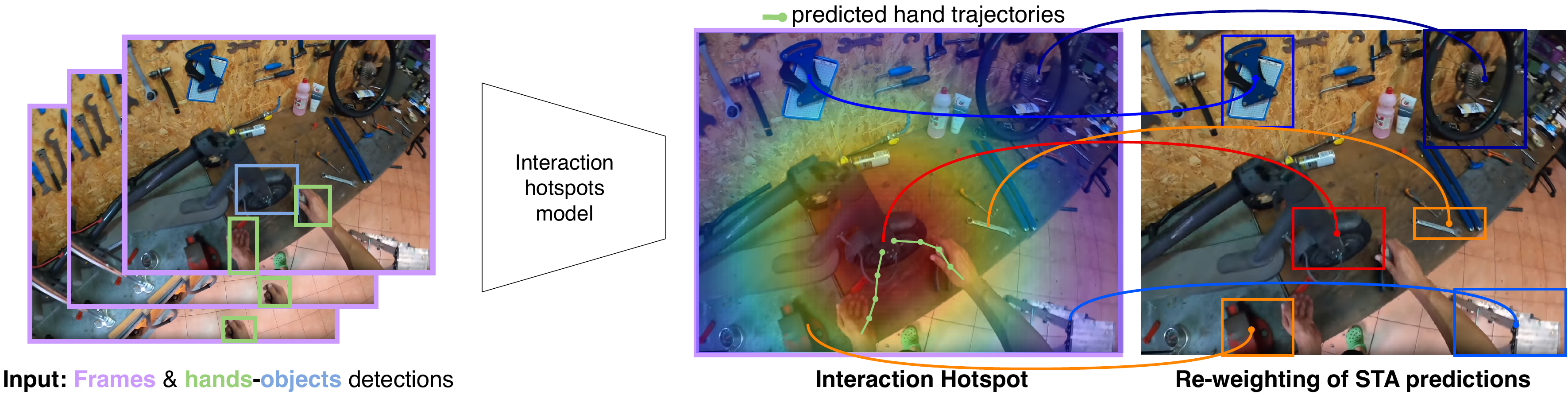}
\caption{\textbf{Refinement of confidence scores based on the interaction hotspots.} The interaction hotspot model observes frames, hands, and objects and forecasts a map encoding the probability of the interaction in each pixel. STA confidence scores are re-weighted based on the probability values at the bounding box coordinate centers, reducing confidence in false positive predictions falling far from the interaction hotspot.}
\label{fig:int_hots_teaset}
\end{figure*}

\subsection{Learning of environment affordances}


The approach described in Section \ref{subsec:fixed_aff} uses affordances to refine predictions at inference time. 
To further improve the exploitation of affordances, here we propose an alternative approach that learns the affordances directly during the training of the end-to-end STA prediction model.
Specifically, we adopt an attention mechanism between the input video descriptor $\Phi(\mathcal{V}')$ and the descriptors $\Phi(\mathcal{Z}_j^\mathcal{V})$ of the affordances zones\textcolor{red}{, where $j$ indexes the different zones}. We interpret the attention mechanism \cite{vaswani2017attention} as a learnable way of querying the most similar situations from the agent's past memory. 
In order to obtain the affordance keys $K_j^{AFF}$, we project with a linear layer $W_K$ the zone video embeddings $\mathcal{Z}_j^\mathcal{V}=  \sum_{i = 1}^{|Z|} \mathrm{\Psi}_\mathcal{V}(\mathcal{V}_i^Z) / |Z|$, \textcolor{red}{where $i$ indexes videos within the zone and $\mathcal{Z}_j$ } represent a descriptor of the environment in zone $Z_j$. In this case, the per-zone video embedding $\mathrm{\Psi}^\mathcal{V}(\mathcal{V}_i^Z)$ is the mean EgoVideo\cite{pei2024egovideo} class token $\overline{C}_{\mathcal{V}}$ of the videos inside the affordance zone $Z_j$.
We represent the affordance query $Q^{AFF}$ by processing the EgoVideo class token of the novel video $C_\mathcal{V'}$ with a learnable linear layer $W_Q$, while keys are computed from the affordance database with a linear layer $W_K$. 
The $W_K, W_Q$ layers learn to compute the similarity of a novel video with respect to all the past environment observations. 
In contrast with a rigid similarity, here we learn how to best associate the input video to the learned affordance zones. \textcolor{red}{Given video $\mathcal{V'}$, we compute a single query $Q^{AFF}$ and a set of keys $K_j^{AFF}$ (one per zone) as follows:}

\begin{equation}
    \textcolor{red}{Q^{AFF} = C_\mathcal{V'} \cdot W_Q
    \quad
    K_j^{AFF} =  {\mathcal{Z}_j}^\mathcal{V} \cdot W_K}
\end{equation}
\noindent

    \quad


\noindent
\textcolor{red}{We then apply a Sigmoid function to the dot product between queries and keys to obtain a similarity scores $S_j$ for each zone $j$,  defined as follows:}

\begin{equation}
    \textcolor{red}{S_j =\text{Sigmoid} (Q^{AFF} \cdot (K_j^{AFF})^T)}
\end{equation}

\noindent

Next, we multiply the per-zone similarity scores $S_j$ by the non-normalized affordance distributions within the zone $Z_j$, defined as $\mathds{1}$ if the noun $\mathcal{N}$ or action verb $\mathcal{A}$ is present, or zero otherwise. We apply a max-pooling operation to obtain the final affordance distribution. As we require per-class binary probabilities due to the Binary Cross Entropy loss adopted from the DETR base model \cite{zhang2022dino}, the max-pooling operation is independent to each class and makes the distribution less sensitive to the long-tail distribution of verbs ad nouns.
We show this computation in the following formula for two individual classes:


\begin{equation}
\begin{aligned}
\textcolor{red}{
p_{\text{aff}}(n=\texttt{cup}|\mathcal{V}')
= \max_{j} \{ S_j \cdot \mathds{1}_{\texttt{cup} \in \mathcal{N}^{Z_j}} \} 
} \\
\textcolor{red}{
p_{\text{aff}}(n=\texttt{take}|\mathcal{V}')
= \max_{j} \{ S_j \cdot \mathds{1}_{\texttt{take} \in \mathcal{A}^{Z_j}} \}
}
\end{aligned}
\end{equation}

\noindent

\noindent
\textbf{Training:} 
\textcolor{red}{We add the nouns affordance distribution $p_{\text{aff}}(n|\mathcal{V'})$ after the final classification layer in the logits space as follows $\log(p_{\text{aff}} + \epsilon) - \log(1 - p_{\text{aff}} + \epsilon)$, which is the input of the Binary Cross Entropy loss for training the model. In this way, we ground the learning of the model on past human behavior, contributing to the full model training. We do the same for verbs.}

\noindent
\textbf{Inference}
\textcolor{red}{During inference, we also add the nouns affordance distribution $p_{\text{aff}}(n|\mathcal{V'})$ transformed to the logits space after the final classification layer. Then, we apply a Sigmoid layer to obtain the final noun binary probabilities. We do the same for verbs.}
In this more flexible approach, the model adapts dynamically to each novel video $\mathcal{V}'$, as we do not rely on a fixed distance or number of zones, learning to attend a memory available at test time.


\section{Leveraging interaction hotspots}
While our affordance database gives us information on which objects (nouns) and interaction modes (verbs) are likely to appear in the current scene, it does not give us any information on \textit{where} the interaction will take place in the observed images.
As noted in previous works~\cite{liu2020forecasting,liu2022joint}, observing how hands move in egocentric videos can allow us to predict the interaction hotspot~\cite{liu2022joint,nagarajan2020ego}, a distribution over image regions indicating possible future interactions locations.
We exploit this concept and include a module to predict an interaction hotspot by observing frames, hands, and objects. As Figure~\ref{fig:int_hots_teaset} illustrates, we hence re-weigh the confidence scores $s_i$ of STA predictions according to the location of the respective bounding box centers in the predicted interaction hotspot, to reduce the influence of false positive detections falling in areas of unlikely interaction.

\vspace{1mm}
\noindent
\subsection{Inferring interaction hotspots} 
We base our interaction hotspot module on the work presented in ~\cite{liu2022joint} with some improvements.
First, we fine-tune the hand object detector presented in~\cite{shan2020understanding} on EGO4D-SCOD~\cite{grauman2022ego4d} annotations, rather than using it out-of-the-box.
Second, we extract stronger egocentric-aware frame features with the video part of the dual-encoder version of EgoVLP~\cite{pramanick2023egovlpv2} pre-trained on Ego4D~\cite{pramanick2023egovlpv2}, instead of using a ConvNet as in~\cite{liu2022joint}.\footnote{See supp. for more information on the interaction hotspot prediction module.} The model takes as inputs the features of the observed frames, besides the coordinates and features of both hands and pre-detected objects, and is trained to forecast the hand trajectory, from which it predicts a distribution over plausible future contact points.
Given the observed image-video pair $(I_T,\mathcal{V}_{T-t:T})$, the output of the model is a probability distribution over the spatial locations of $I_T$ indicating the probability of interaction of each pixel denoted as $p_{ih}(x, y |  I_T, \mathcal{V}_{T - t:T})$. 


\vspace{1mm}
\noindent
\subsection{Fusing STA predictions with interaction hotspots:}
We exploit the interaction hotspots to refine the predictions of the STA model, assuming that regions close to the predicted interaction hotspots are more likely to contain the next active objects. 
Given a predicted box $\hat b_i$, we re-weigh its related confidence score $\hat s_i$ according to the location of the bounding box center $(\hat c_i^x, \hat c_i^y)$ in the interaction hotspot as following: $\hat s_i \cdot p_{ih}(\hat c_i^x, \hat c_i^y|I_T, \mathcal{V}_{T - t:T})$. 

\begin{table*}[t]
    \centering
    \begin{minipage}{0.49\linewidth}
    \centering
            \caption{Results in mAP on the validation split of Ego4D-STA v1. \textbf{Best results} in bold. Relative gain is with respect to \underline{second best}}
            \label{tab:mAP_v1}
        \begin{tabular}{|l|cccc|}
        \hline
        Model & N & N + V & N + $\delta$ & All \\ \hline
        FRCNN+SF~\cite{grauman2022ego4d} & 17.55 & 5.19 & 5.37 & 2.07 \\
        FRCNN+Feat.~\cite{sta2023quickstart} & 22.01 & 5.52 & 5.54 & 1.78 \\
        StillFast~\cite{ragusa2023stillfast}  & 16.21 & 7.47 & 4.94 & 2.48 \\
        Transfusion~\cite{pasca2023summarize} & 20.19 & 7.55 & 6.17 & 2.60 \\ \hline
        STAformer & 21.71 & 10.75 & 7.24 & 3.53 \\
        STAformer \& AFF (fixed)& \underline{24.36} & \underline{12.00} & \underline{7.66}  & \underline{3.77} \\ \hline
        STAformer++ & 32.07 & 15.00 & 8.53 & 4.31 \\
        STAformer++ \& AFF (learned) & \textbf{33.21} & \textbf{15.94} & \textbf{8.98} & \textbf{4.66} \\
        \hline
        Gain (rel $\%$) & \textcolor{ForestGreen}{+36.3} & \textcolor{ForestGreen}{+24.5} & \textcolor{ForestGreen}{+17.2} & \textcolor{ForestGreen}{+23.6}  \\ \hline
        \end{tabular}
    \end{minipage}
    \hfill
    \begin{minipage}{0.47\linewidth}
    \centering
            \caption{Results in mAP on the validation split of Ego4D-STA v2.  \textbf{Best results} in bold. Relative gain is with respect to \underline{second best}.}
            \label{tab:mAP_v2}
        \begin{tabular}{|l|cccc|}
            \hline
            Model & N & N + V & N +$\delta$ & All \\ \hline
            FRCNN+SF~\cite{grauman2022ego4d} & 21.00 & 7,45 & 7.07 & 2.98 \\
            InternVideo~\cite{chen2022internvideo} & 19.45 & 8.00 & 6.97 & 3.25 \\
            StillFast~\cite{ragusa2023stillfast} & 20.26 & 10.37 & 7.26 & 3.96 \\
            GANO v2~\cite{thakur2023guided} & 20.52 & 10.42 & 7.28 & 3.99 \\ \hline
            STAformer & 27.51 & 14.68 & 9.63 & 5.50 \\
            STAformer \& AFF (fixed) & \underline{29.39} & \underline{15.38} & \underline{9.94} & \underline{5.67} \\ \hline
            STAformer++ & 36.78 & 17.26 & 11.03 & 5.87 \\
            STAformer++ \& AFF (learned) & \textbf{37.41} & \textbf{18.51} & \textbf{11.14} & \textbf{6.26} \\
            \hline
            Gain (rel $\%$) & \textcolor{ForestGreen}{+27.8} & \textcolor{ForestGreen}{+20.3} & \textcolor{ForestGreen}{+12.7} &  \textcolor{ForestGreen}{+10.4} \\ \hline
        \end{tabular}
    \end{minipage}
\end{table*}

\section{Experimental setup}

Following the official benchmark \cite{grauman2022ego4d}, we adopt standard Noun (N), Noun+Verb (N+V), Noun+time-to-contact (N+$\delta$) and Noun+Verb+time-to-contact (All) Top-5 mean Average Precision (mAP). We also provide detailed comparative on the Top-5 Average Precision metric (AP) as defined in \cite{thakur2023guided}. 

\subsection{Datasets}

We validate our method on Ego4D~\cite{grauman2022ego4d} and EPIC-Kitchens~\cite{damen2018scaling}, two large-scale datasets of egocentric videos with high diversity and long-tail distributions classes.

\textbf{Ego4D.} We consider both the first and second versions of Ego4D STA annotations. Version 1 (v1) of the Ego4D STA split is composed of 27,801 training 17,217 validation and 19,870 test instances with 87 noun and 74 verb categories. Version 2 (v2) extends v1 with additional videos and annotations, for a total of 98,276 training, 47,385 validation and 19,870 test videos, with a taxonomy of 128 nouns and 81 verb classes. Ego4D STA contains a single test split which is compatible with v1 and v2, hence models trained on either versions can be compared on the same test split.

\textbf{EPIC-Kitchens.}
Since Ego4D is the only dataset containing STA annotations to date, we extend EPIC-Kitchens dataset~\cite{damen2018scaling} by post-processing its active object and action segment annotations.
We first merge active object bounding boxes into tracks by grouping neighboring annotations of the same object class and removing tracks with multiple bounding boxes for the same class.
Then, we match each object track to one of the annotated action segments. Specifically, if an action segment including the same noun as the object track is found, this is matched to the object track.
Next, we truncate object tracks to exclude frames depicting the action, enabling anticipation of future actions.
Finally, we attach the following data to a given bounding box: the noun associated to the track as the object category, the associated action segment as the interaction verb, the distance from the time-step of the current frame to the beginning of the associated action segment as the time to contact.
The final set of annotations contains 33,804 training and 7,055 validation instances with 104 noun and 51 verb categories, which we \href{https://github.com/lmur98/AFFttention}{release} to the community.

\subsection{Implementation details}

We train STAformer with Adam as an optimizer, an initial learning rate of $10^{-4}$ with linear warm-up, and a weight decay of $10^{-6}$, on 4 Tesla V100 GPUs. Similarly, we train STAformer++ following the official procedure of DINO-DETR\cite{zhang2022dino} with AdamW as optimizer and an initial learning rate of $2\cdot10^{-5}$.  
For the image encoders, we first adopt the DINOv2-B \cite{oquab2023dinov2} visual transformer, composed by 12 blocks. Alternatively, we extract multi-scale image features with the Swin-T \cite{liu2021swin} Large version pre-trained on COCO dataset formed by 24 blocks grouped in hierarchical depths. We fine-tune the last 3 blocks in both cases. 
For the video encoders, we initialize the TimeSformer weights with the dual encoder version of EgoVLP-v2 \cite{pramanick2023egovlpv2}, formed by 12 blocks. 
We utilize the video encoder of EgoVIDEO\cite{pei2024egovideo} composed of 38 blocks. 
For TimeSformer, we sample 16 frames at 30 FPS, while the available version of EgoVideo only allows processing 4 frames, which we sample at 7.5 FPS to cover the same video segment.
In both cases, we fine-tune the last 4 blocks of the video model. The DETR model weights are initialized using the 12-epoch version of \cite{zhang2022dino}.

\section{Results}


We compare our model against several STA baselines which either provide open source implementations~\cite{grauman2022ego4d,ragusa2023stillfast} or report results in their papers\cite{grauman2022ego4d, sta2023quickstart, ragusa2023stillfast, pasca2023summarize, chen2022internvideo, thakur2023guided}. We also report multiple ablation studies showing the contribution of each individual component of our approach.

\subsection{Comparison with the state-of-the-art}

\noindent
\textbf{Ego4D v1 validation split (Tables \ref{tab:mAP_v1}-\ref{tab:AP_metrics}).}
Our initial version of STAformer achieves 21.71 N, 10.75 N+V, 7,24 N+$\delta$ and 3.53 All mAP, while our the most advanced version of STAformer plus the incorporation of learned affordances, scores a 33.21 N, 15.94 N+V, 8.89 N+$\delta$ and 4.88 All mAP in the v1 split, as Table~\ref{tab:mAP_v1} shows.
We obtain a relative gain\footnote{{We compute the relative gain\% of $x$ relative to $y$ as $100 \cdot (\frac{x-y}{y})$.}} up to +23.6 $\%$ in the mAP All metric compared with our previous conference version \cite{mur2025aff}. 
Table \ref{tab:AP_metrics} compares our method with previous approaches reporting results using the AP metric.
In this case, the initial version of STAformer, based on a Faster-RCNN architecture, shows a lower detection performance (38.38 B AP) compared with the most advanced version of GANO \cite{thakur2024leveraging} (45.30 Box AP). However, the novel DETR-based architecture of STAformer++ notably improves the quality of the predicted bounding boxes up to 49.68 B AP (+9.7 $\%$ relative gain), which is reflected in the overall metric where it achieves a 4.77 (+ 17.5 $\%$ relative gain).
The results also show the benefits of leveraging affordances in the short-term anticipation task, which are specially relevant in the semantic metrics (+ 37.9 $\%$ B+N AP, +16.7 $\%$ B+V AP, +24.5 $\%$ B+N+V AP, + 36.3 $\%$ N mAP, + 24.5 $\%$ N+V mAP), since this prior just refines the noun and verb probabilities.

\noindent
\textbf{Ego4D v2 validation split (Tables \ref{tab:mAP_v2}-\ref{tab:AP_metrics}).}
The overall improvement is also significant in the v2 split, a larger version of Ego4D containing also v1 annotations. 
Our most advanced version, STAformer++ with learned affordances, scores 37.41 N mAP, 18.51 N+V mAP, 11.14 N+$\delta$ and 6.26 All mAP, showing a relative gain of +10.4 $\%$ in the overall metric and demonstrating significant improvements both in semantic, detection, and temporal performance.
%
%


\begin{table*}[]
\centering
\caption{Results in AP metric on the validation split of Ego4D-STA.  \textbf{Best results} in bold. Relative gain is with respect to \underline{second best}. T denotes training data.}
\label{tab:AP_metrics}
\begin{tabular}{|l|ccccccccc|}
\hline
 & T & B & B+N & B+V & B+$\delta$ & B+N+V & B+N+$\delta$ & B+V+$\delta$ & All \\ \hline
Slowfast & v1 & 40.50 & 24.50& 0.34& 8.16& 0.34& 5.00& 0.06 & 0.06 \\
Slowfast (w/Transformer) & v1 & 40.50 & 24.50& 8.20& 7.50& 8.20& 4.50& 1.30& 0.73\\
AVT & v1 & 40.50 & 24.50 & 8.45& 7.12& 8.45& 4.39& 1.15& 0.71\\
ANACTO & v1 & 40.50 & 24.50 & 8.90& 7.47& 8.90& 4.55& 1.54& 0.91\\
MeMVIT & v1 & 40.50 & 24.50 & 10.05& 9.27& 10.04& 4.95& 2.11& 1.34\\
GANO (w/o guided attention) & v1 & 40.50 & 24.50 & 7.10& 9.01& 7.10& 4.20& 1.22& 0.75\\
GANO (w/ guided attention) & v1 & \underline{45.30} & 25.80& 10.56& 10.1& 10.56& 5.90& 2.77& 1.70\\
StillFast & v1 &  27.78&  17.75&  10.21&  7.33&  7.01&  4.61&  2.68&  1.77\\ \hline
STAformer & v1 &  38.38&  \underline{28.36}&  \underline{16.66}&  \underline{12.27}&  \underline{12.66}&  \underline{8.89}&  \underline{5.47}&  \underline{4.06}\\ \hline
STAformer++ & v1 &  \textbf{49.68} &  38.83&  18.02 &  \textbf{12.92}&  14.00 & 10.58 & \textbf{5.75} &  4.63\\ 
STAformer++ \& AFF(learned) & v1 & 48.60& \textbf{39.12} & \textbf{19.45} & 12.29& \textbf{15.77} & \textbf{10.28}& 5.71& \textbf{4.77}\\ \hline
Gain & & \textcolor{ForestGreen}{+9.7}& \textcolor{ForestGreen}{+37.9}& \textcolor{ForestGreen}{+16.7}& \textcolor{ForestGreen}{+5.3}& \textcolor{ForestGreen}{+24.5}& \textcolor{ForestGreen}{+15.6}& \textcolor{ForestGreen}{+4.4}& \textcolor{ForestGreen}{+17.5}\\ \hline \hline
STAformer & v2 & \underline{43.24}& \underline{33.53}& \underline{20.88}& \underline{14.84}& \underline{16.52}& \underline{11.23}& \underline{7.70}& \underline{5.89}\\ \hline
STAformer++ & v2 & 55.95& 47.02& 24.40& \textbf{16.91}& 20.24& \textbf{14.14}& 8.40& \textbf{6.85}\\ 
STAformer++ \& AFF(learned) & v2 & \textbf{55.98}& \textbf{47.63}& \textbf{24.82}& 16.40& \textbf{20.77}& 13.86& \textbf{8.42}& 6.77\\ \hline
Gain & & \textcolor{ForestGreen}{+29.3}& \textcolor{ForestGreen}{+42.2}& \textcolor{ForestGreen}{+18.9}& \textcolor{ForestGreen}{+13.9}& \textcolor{ForestGreen}{+25.7}& \textcolor{ForestGreen}{+25.9}& \textcolor{ForestGreen}{+9.3} & \textcolor{ForestGreen}{+14.9}\\
\hline
\end{tabular}%
\end{table*}

\begin{table}[t]
    \centering
        \caption{Results in mAP on the test split of Ego4D-STA of models trained on the v1 training split.}
        \label{tab:mAP_test_v1}
        \begin{tabular}{|l|cccc|}
        \hline
        Model  & N & N + V & N + $\delta$ & All \\ \hline
        FRCNN+SF.~\cite{grauman2022ego4d} & 20.45 & 6.78 & 6.17 & 2.45\\
        FRCNN+Feat.~\cite{sta2023quickstart} & 20.45 & 4.81 & 4.40 & 1.31 \\
         InternVideo ~\cite{chen2022internvideo}& 24.60& 9.18& 7.64 &3.40\\
         Transfusion~\cite{pasca2023summarize}& 24.69& 9.97& 7.33&3.44\\
         StillFast \cite{ragusa2023stillfast} & 19.51& 9.95& 6.45& 3.49\\ \hline
         STAformer & 24.39 & 12.49& 7.54&4.03\\ 
         STAformer \& AFF (fixed) & \underline{26.52}& \underline{13.15}& \underline{7.78}& \underline{4.06}\\ \hline
         STAformer++ & 33.78 & 14.28 & \textbf{10.14} & 4.97  \\
         STAformer++\&AFF(learned)& \textbf{34.06} & \textbf{15.94} & 10.10 & \textbf{5.24}  \\ \hline 
         Gain (rel $\%$) & \textcolor{ForestGreen}{+28.4} & \textcolor{ForestGreen}{+21.2} & \textcolor{ForestGreen}{+29.8} & \textcolor{ForestGreen}{+29.1} \\ \hline 
         \end{tabular}
\end{table}

\begin{table}[t]
    \centering
        \caption{Results in mAP on the test split of Ego4D-STA of models trained on the v2 training split.}
        \label{tab:mAP_test_v2}
        \begin{tabular}{|l|cccc|}
        \hline
        Model  & N & N + V & N + $\delta$ & All \\ \hline
        StillFast \cite{ragusa2023stillfast}   & 25.06 & 13.29 & 9.14 & 5.12 \\
        GANO v2 \cite{thakur2023guided}  & 25.67 & 13.60 & 9.02 & 5.16 \\
        Language NAO  & 30.43 & 13.45 & 10.38 & 5.18 \\ 
        EgoVideo & 31.08 & 16.18 & \underline{12.41} & \textbf{7.21} \\ \hline
        STAformer & 30.61 & 16.67 & 10.06 & 5.62 \\
        STAformer \& AFF (fixed) & \underline{32.39} & \underline{17.38} & 10.26 & 5.70 \\ \hline
        STAformer++ & 41.96 & 19.16 & \textbf{13.05} & \underline{6.92} \\ 
        STAformer++\&AFF(learned)& \textbf{42.07} & \textbf{19.51} & 12.73 & 6.26 \\ \hline
        Gain (rel $\%$) & \textcolor{ForestGreen}{+29.9} & \textcolor{ForestGreen}{+12.3} & \textcolor{ForestGreen}{+5.2} & \textcolor{orange}{-4.1}  \\ \hline
        \end{tabular}
\end{table}      

\begin{table}[]
    \centering
        \caption{Results in mAP on the validation split of EPIC-Kitchens. \textbf{Best results} in bold. Relative gain is with respect to \underline{second best}.}
        \label{tab:mAP_EK}

        \begin{tabular}{|l|cccc|}
            \hline
            Model & N & N + V & N + $\delta$ & All \\ \hline
            StillFast \cite{ragusa2023stillfast} & 21.24 & 12.41 & 6.22 & 3.28 \\ 
            \hline
            STAformer & 25.25 & 17.17 & 9.10 & 6.13 \\
            STAformer \& AFF (fixed) & \underline{28.37} & \underline{18.95} & \underline{9.29} & \underline{6.60}\\ \hline
            STAformer++ & 44.96 & 24.67 & 14.01 & 7.87 \\
            STAformer++\&AFF(learned)& \textbf{45.34} & \textbf{25.82} & \textbf{14.06} & \textbf{8.67} \\
            \hline
            Gain (rel $\%$) & \textcolor{ForestGreen}{+59.5} & \textcolor{ForestGreen}{+36.20} & \textcolor{ForestGreen}{+51.6} & \textcolor{ForestGreen}{+31.5} \\ \hline
        \end{tabular}  
\end{table}

\noindent
\textbf{Ego4D test split (Table \ref{tab:mAP_test_v1}-\ref{tab:mAP_test_v2}).}
Since the test set of Ego4D is private, we are only able to compare approaches showing test results in their papers. 
For fair comparisons, we report two settings with methods trained on v1 or v2. Our method achieves significant gains with respect to trained methods on v1, for instance, obtaining a $+28.4\%$ N mAP, $+21.2\%$ N+V mAP, + $29.8\%$ N+$\delta$ mAP and $+29.1\%$ in mAP All.
We observe similar improvements when training on v2, with $+29.9\%$ N mAP, $+12.3\%$ N+V mAP and $+5.2\%$ N+$\delta$ mAP. However, our model does not outperform EgoVideo in the overall metric, scoring 6.92 vs. 7.21 All mAP. Note that the participating version of EgoVideo is fully fine-tuned to the STA task on Ego4D and covers 16 frames, while, for computational constraints, our STAformer++ model just trains the last 4 blocks of a simpler general model which only processes 4 frames.
It is worth noting that our approach also benefits from training on larger datasets, improving from 5.24 mAP All when it is trained on v1 (Table~\ref{tab:mAP_test_v1}) to 6.92 mAP All when training on v2 Table~\ref{tab:mAP_test_v2}.

\noindent
\textbf{EPIC-Kitchens STA (Table \ref{tab:mAP_EK}).}
Since this benchmark is new, we train the official implementation of StillFast~\cite{ragusa2023stillfast} on EPIC-Kitchens as a baseline, obtaining 21.24 N mAP, 12.41 N+V mAP, 6.22 N + $\delta$ mAP and 3.28 All mAP.
The introduction of more powerful backbones (DINO and TimeSformer) and the dual cross-attention mechanism in STAformer achieve a 28.37 N mAP, 18.95 N+V mAP, 9.29 N + $\delta$ mAP and 6.60 All mAP. 
The performance gains are particularly notable with STAformer++, achieving 45.34 N mAP, 25.82 N+V mAP, 14.06 N+$\delta$ mAP and 8.67 All mAP, representing a +31.5 $\%$ increase in All mAP. This highlights the generality of our framework in different training regimes and datasets.

\begin{table*}[t]
\centering
\caption{Ablation study of the architectural components of STA-former on the validation split of Ego4D-v1.  \faSnowflake Encoder frozen \faFire Encoder fine-tuned. \textbf{Overall best results} in bold. \underline{Best result of each part} is underlined. For fair comparison, we fine-tune 3 blocks in the image encoders and 4 blocks in the video encoders and the video comprises 0.5 sec. \textcolor{red}{Configurations using the Faster R-CNN head correspond to STAformer, whereas configurations using the DETR head are instantiations of STAformer++}}
\label{tab:staformer_ablation}

\scriptsize
\resizebox{\textwidth}{!}{%
\begin{tabular}{|c|ccccc|cccc|}
\hline
 Exp.& Image Encoder & Video Encoder & Temporal pooling & 2D-3D Fusion  &Detection Head& N & N + V & N + $\delta$ & All \\ \hline
 \cite{ragusa2023stillfast}&R50 \faCog  & X3D \faCog  & Mean & Sum  &Fast-RCNN& 16.21 & 7.52 & 4.94 & 2.48 \\ \hline \hline
 A1&DINOv2  \faSnowflake & - & - & -  &Fast-RCNN& 17.48 & 8.64 & 5.20 & 2.52 \\
 A2& Swin-T  \faSnowflake & - & - & -  & DETR& 27.69 & 9.57 & 5.43 &2.71 \\
 A3& Swin-T  \faFire    & - & - & -  & DETR& 28.77 & 11.04 & 6.12 &2.85 \\ 
 A4& DINOv2  \faFire & - & - & -  &DETR& \underline{29.33} & \underline{11.65} & \underline{6.46} & \underline{2.98} \\\hline \hline
 
 B1&DINOv2  \faSnowflake & DINOv2  \faSnowflake & Mean & Sum  &Fast-RCNN& 15.82& 7.65 & 4.11 & 2.19 \\
 B2&DINOv2  \faSnowflake & X3D \faFire & Mean& Sum &Fast-RCNN& 18.84 & 8.84 & 5.56 & 2.57 \\
 B3&DINOv2 \faSnowflake & TimeSformer \faFire & Mean & Sum  &Fast-RCNN& 16.67 & 8.38 & 5.16 & 2.63 \\ 
 B4&DINOv2 \faSnowflake & TimeSformer \faFire & Mean & Sum & DETR & 26.11 & 10.65 & 6.90 & 3.02 \\
 B5&Swin-T \faFire & TimeSformer \faFire & Mean & Sum & DETR & 24.55 & 9.49 & 6.01 & 2.89 \\
 B5&Swin-T  \faFire & EgoVideo \faSnowflake & Mean & Sum & DETR & 31.82 & 13.15 & 6.81 & 3.24 \\
 B6&Swin-T  \faFire & EgoVideo \faFire & Mean & Sum & DETR & \underline{32.50} & \underline{14.72} & \underline{7.73} & \underline{3.65} \\ \hline \hline

 C1&DINOv2  \faSnowflake & TimeSformer \faFire & Conv & Sum  &Fast-RCNN& 17.36 & 8.75 & 6.05 & 2.94 \\ 
 C2&DINOv2  \faSnowflake & TimeSformer \faFire & SH.Attn & Sum  &Fast-RCNN& 19.78 & 10.04 & 6.35 & 3.39 \\ 
 C3&Swin-T  \faFire   & EgoVideo \faFire & MH.Attn & Sum & DETR & 31.31 & 14.22 & 8.05 & 4.07 \\ 
 C4&Swin-T  \faFire   & EgoVideo \faFire & per-Scale MH.Attn & Sum & DETR & \textbf{\underline{32.57}} & \textbf{\underline{15.10}} & \textbf{\underline{8.53}} & \underline{4.31} \\ 
 C5&DINOv2  \faFire   & EgoVideo \faFire & MH.Attn & Sum & DETR & 31.15 & 13.72 & 7.71 & 3.84 \\
 C6&DINOv2  \faFire   & EgoVideo \faFire & per-Scale MH.Attn & Sum & DETR & 31.73 & 14.10 & 8.21 & 4.26 \\ \hline \hline

  D1&DINOv2  \faSnowflake& TimeSformer \faFire & SH.Attn & Dual $I \leftrightarrow \mathcal{V}$ attn  &Fast-RCNN& 20.08& 10.21& 6.51& 3.47 \\
  D2&DINOv2  \faFire & TimeSformer \faFire & SH.Attn & Dual $I \leftrightarrow \mathcal{V}$ Attn &Fast-RCNN& 21.71& 10.75 & 7.24& 3.53 \\
  D3&DINOv2  \faFire & TimeSformer \faFire & SH.Attn & $I\xrightarrow{}\mathcal{V} $ Attn  &Fast-RCNN& 20.01 & 10.04 & 5.80 & 3.01\\
  D4&DINOv2  \faFire & TimeSformer \faFire & SH.Attn & $\mathcal{V}\xrightarrow{}I$ Attn &Fast-RCNN& 20.12& 10.31& 6.30& 3.35\\
  D5&DINOv2  \faFire & TimeSformer \faFire & MH.Attn & MH.Dual $I\leftrightarrow\mathcal{V}$ Attn &Fast-RCNN& 23.02 & 11.57& 7.86& 3.85 \\ 
  D6&Swin-T  \faFire   & EgoVideo \faFire & per-Scale MH.Attn & MH.Dual $I \leftrightarrow \mathcal{V}$ Attn& DETR & 31.80 & \underline{15.06} & 7.95 & \textbf{\underline{4.54}} \\ 
  D7&DINOv2  \faFire   & EgoVideo \faFire & per-Scale MH.Attn & MH.Dual $I \leftrightarrow \mathcal{V}$ Attn& DETR & \underline{31.82} & 13.92 & \underline{7.98} & 4.21 \\ \hline
  
\end{tabular}%
}
\end{table*}

\subsection{Ablation Study on STAformer and STAformer++ components}

Table~\ref{tab:staformer_ablation} ablates the performance effect of the proposed components of the STA models: the image encoder, the video encoder, the temporal pooling, the 2D-3D fusion module and the prediction head (Faster-RCNN or DETR based).

\noindent
\textbf{Image encoder and STA head (Table~\ref{tab:staformer_ablation}, Exp A).} 
We first encode the image with DINOv2 (Exp A.1) and discard the video, obtaining small gains with respect to the baseline~\cite{ragusa2023stillfast}.
While \cite{ragusa2023stillfast} fully trains both image-video encoders, the A1 version trains solely the Faster-RCNN STA prediction head and reflects the modeling capacity of DINOv2.
Then, we replace the Faster-RCNN head by the DETR\cite{zhang2022dino} in Experiments A2-A4. When the entire Swin-T is frozen and uses existing weights pre-trained on COCO(Exp. A2), it losses the generalization capabilities of DINOv2 features, obtaining a drop to 1.91 mAP in the All metric. Then, when we refine the last blocks of both image encoders the performance increases, achieving a 2.55 mAP All for the Swin-T version and a 2.88 mAP All for the DINOv2 model. 
The main benefit of using a DETR-based model is its superior detection capabilities, as the Noun mAP shows an improvement from 17.48 (DINOv2 with Faster-RCNN) to 29.33 (DINOv2 with DETR).

\noindent
\textbf{Video encoder (Table~\ref{tab:staformer_ablation}, Exp B).} 
Using per-frame DINOv2 features with mean temporal pooling (Exp. B1 vs. A1) reduces performance, highlighting DINOv2's limitations in capturing video dynamics. 
However, incorporating an specific video encoder like the X3D 3D CNN \cite{feichtenhofer2019slowfast} (Exp. B2 vs. A1) achieves better results, 
indicating the advantage of appropriately encoding video dynamics.
Experiments B3, B4 and B5 show that adopting TimeSformer as video model only leads to marginal improvements with respect to A4, B2 and A3, respectively.
Incorporating EgoVideo improves the semantic and temporal reasoning. 
Indeed, finetunning the last blocks of EgoVideo (Exp. B6) achieves a 7.73 N+$\delta$ mAP, due to the direct connection of the EgoVideo class token $C_\mathcal{V}$ with the temporal MLP.


\begin{figure}[ht]
    \centering
    \includegraphics[width=0.24\textwidth]{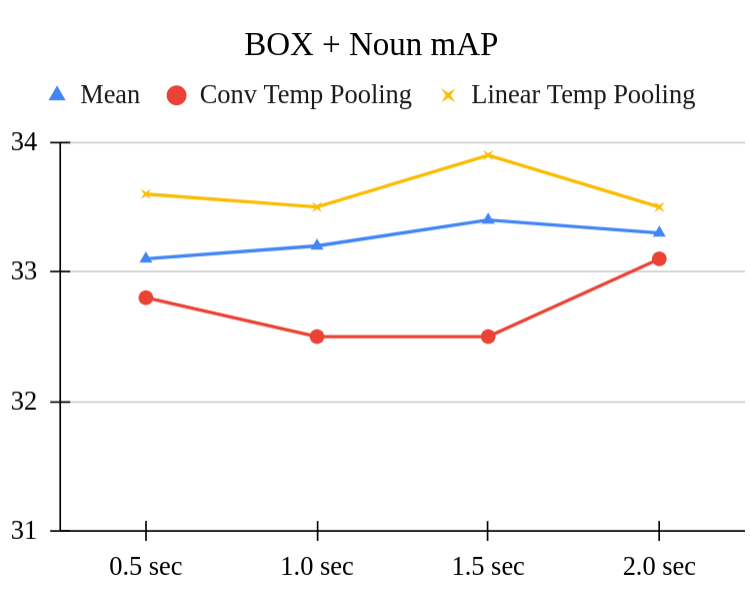}
    \includegraphics[width=0.24\textwidth]{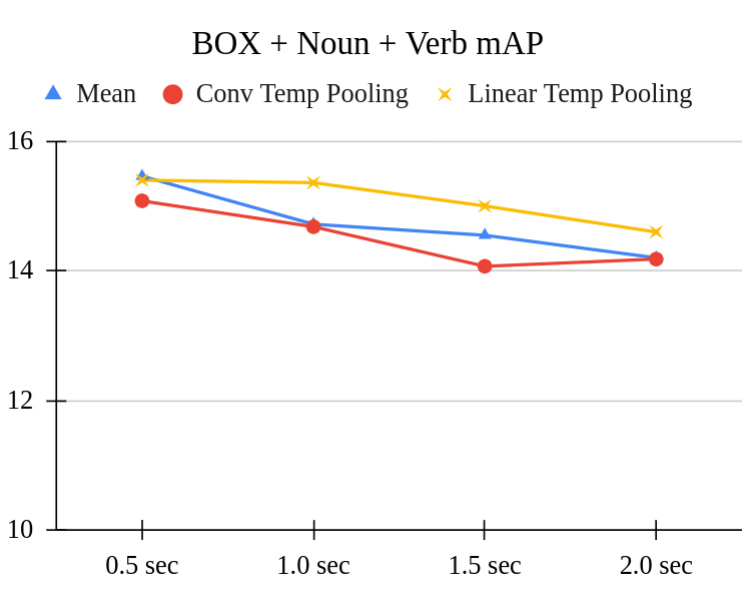} \\
    \includegraphics[width=0.24\textwidth]{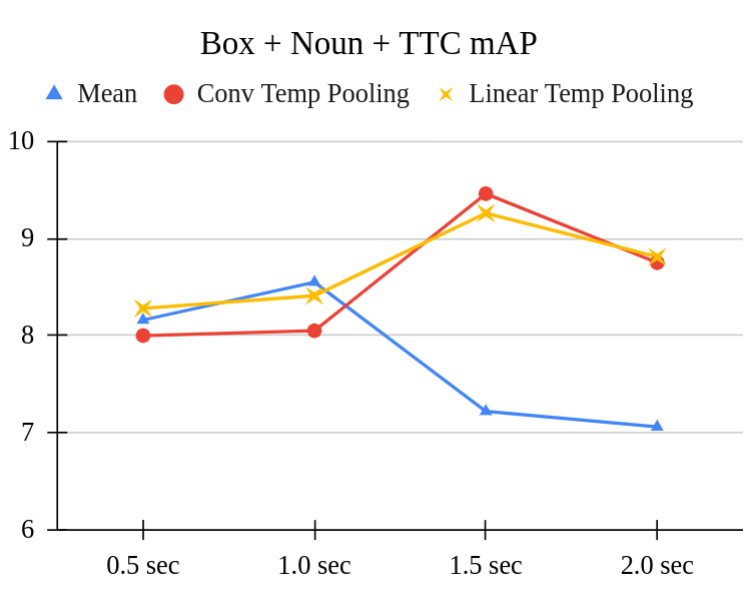}
    \includegraphics[width=0.24\textwidth]{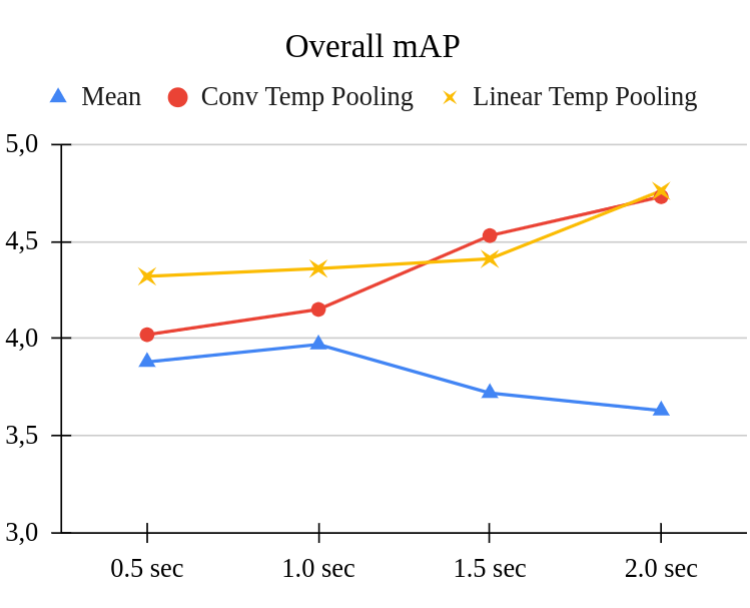}
   
    \caption{\textbf{STAformer++ performance evolution according to the amount of video seen.} We report the mAP N, mAP N+V, mAP N+$\delta$, mAP Overall on the validation split of Ego4D-STA v1.}
    \label{fig:time_temp_pool}
\end{figure}

\begin{figure}[ht]
    \centering
    \includegraphics[width=0.24\textwidth]{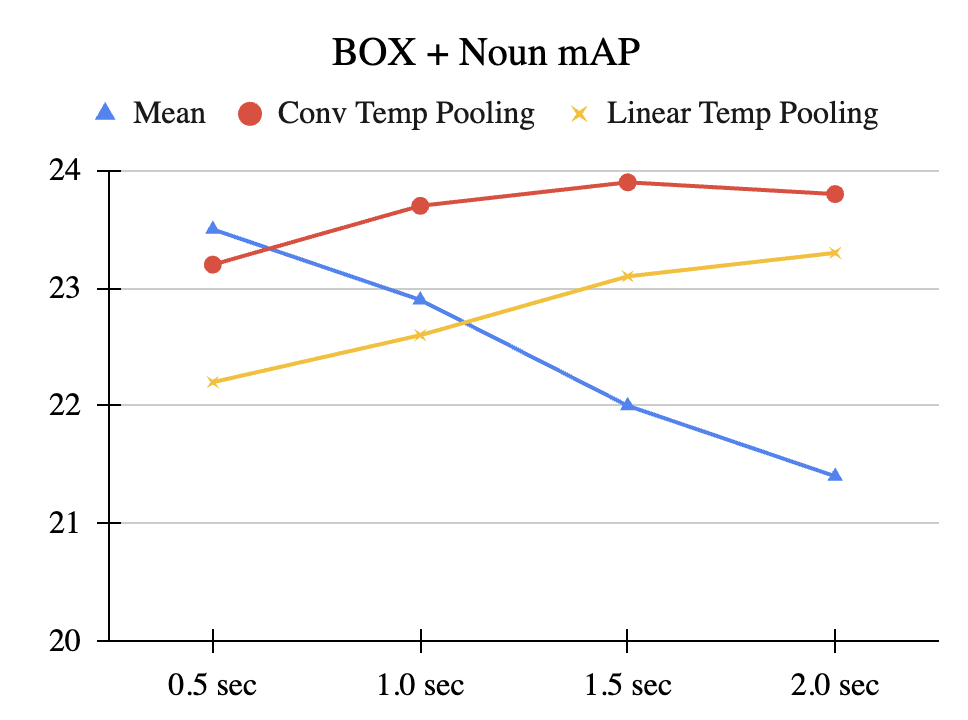}
    \includegraphics[width=0.24\textwidth]{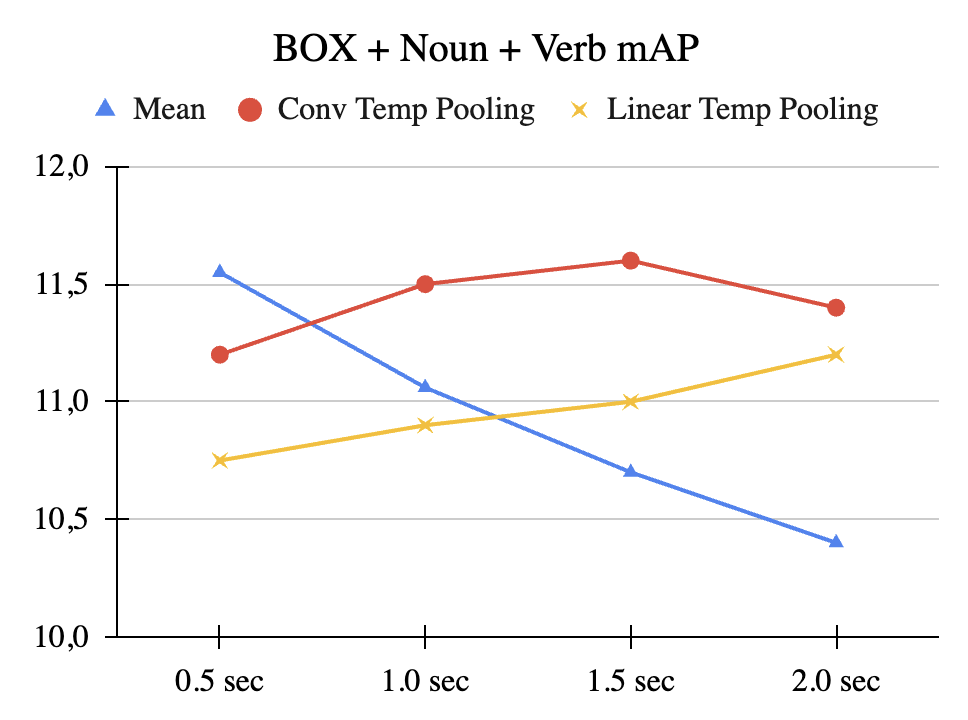} \\
    \includegraphics[width=0.24\textwidth]{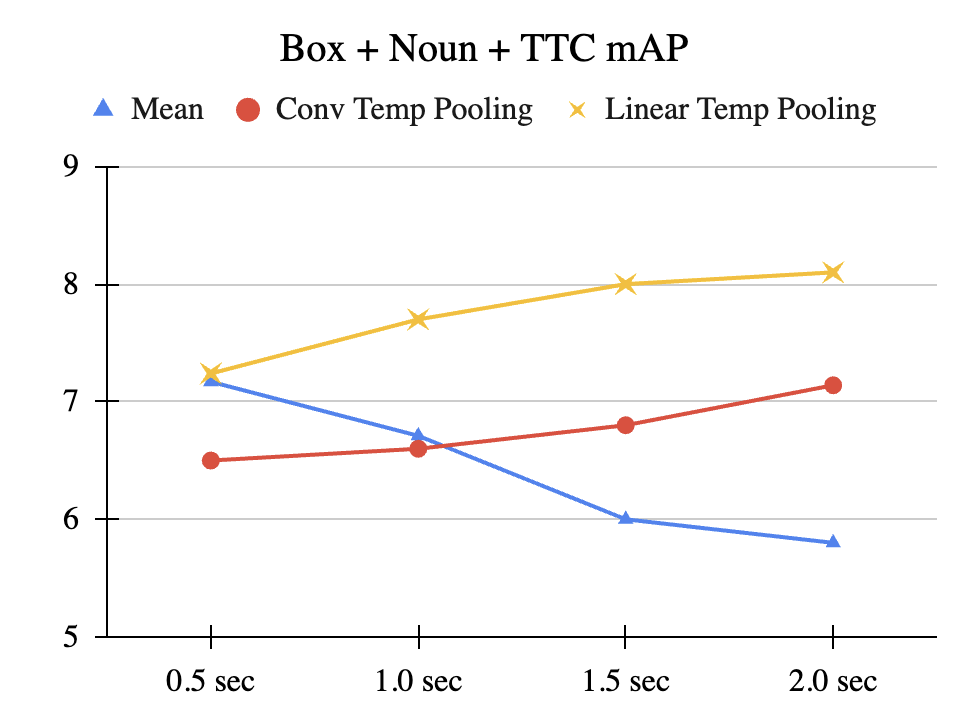}
    \includegraphics[width=0.24\textwidth]{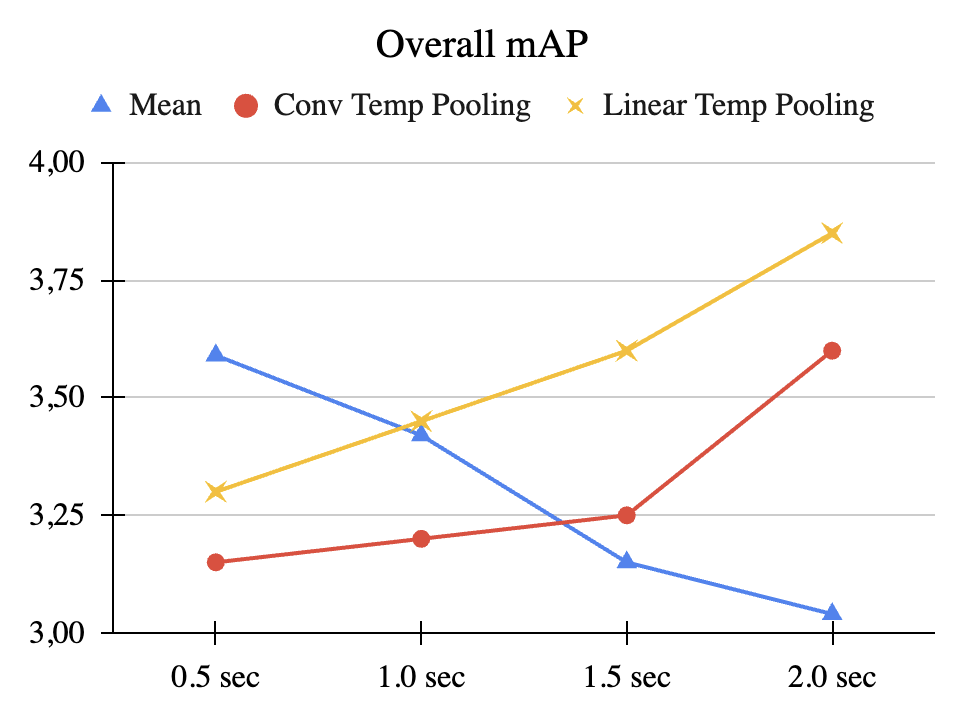}
   
    \caption{\textbf{\textcolor{red}{STAformer performance evolution according to the amount of video seen.}} We report the mAP N, mAP N+V, mAP N+$\delta$, mAP Overall on the validation split of Ego4D-STA v1.}
    \label{fig:time_temp_pool_staformer_original}
\end{figure}

\begin{figure*}[ht]
\centering
\includegraphics[width=0.255\textwidth]{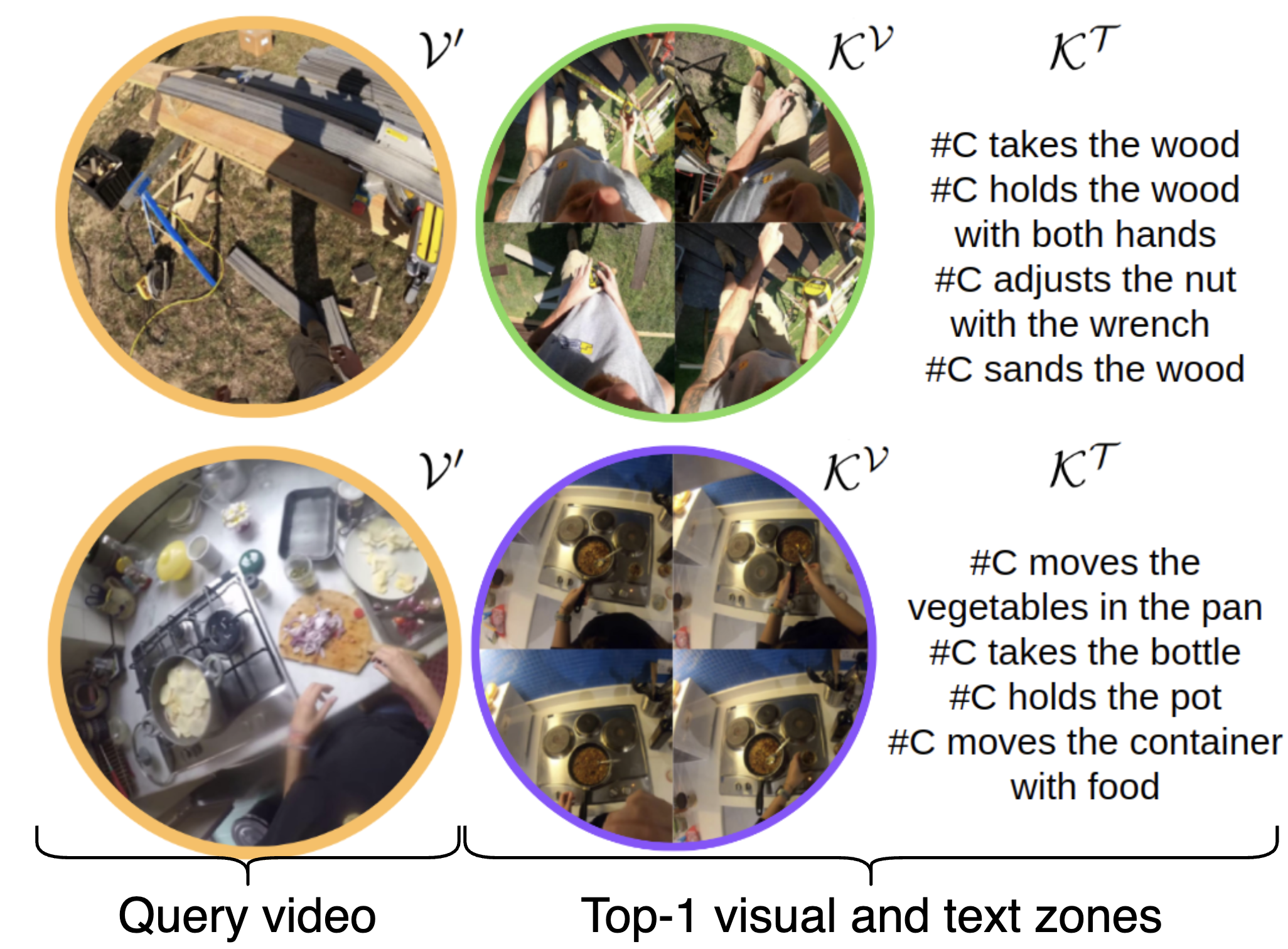} 
\includegraphics[width=0.37\textwidth]{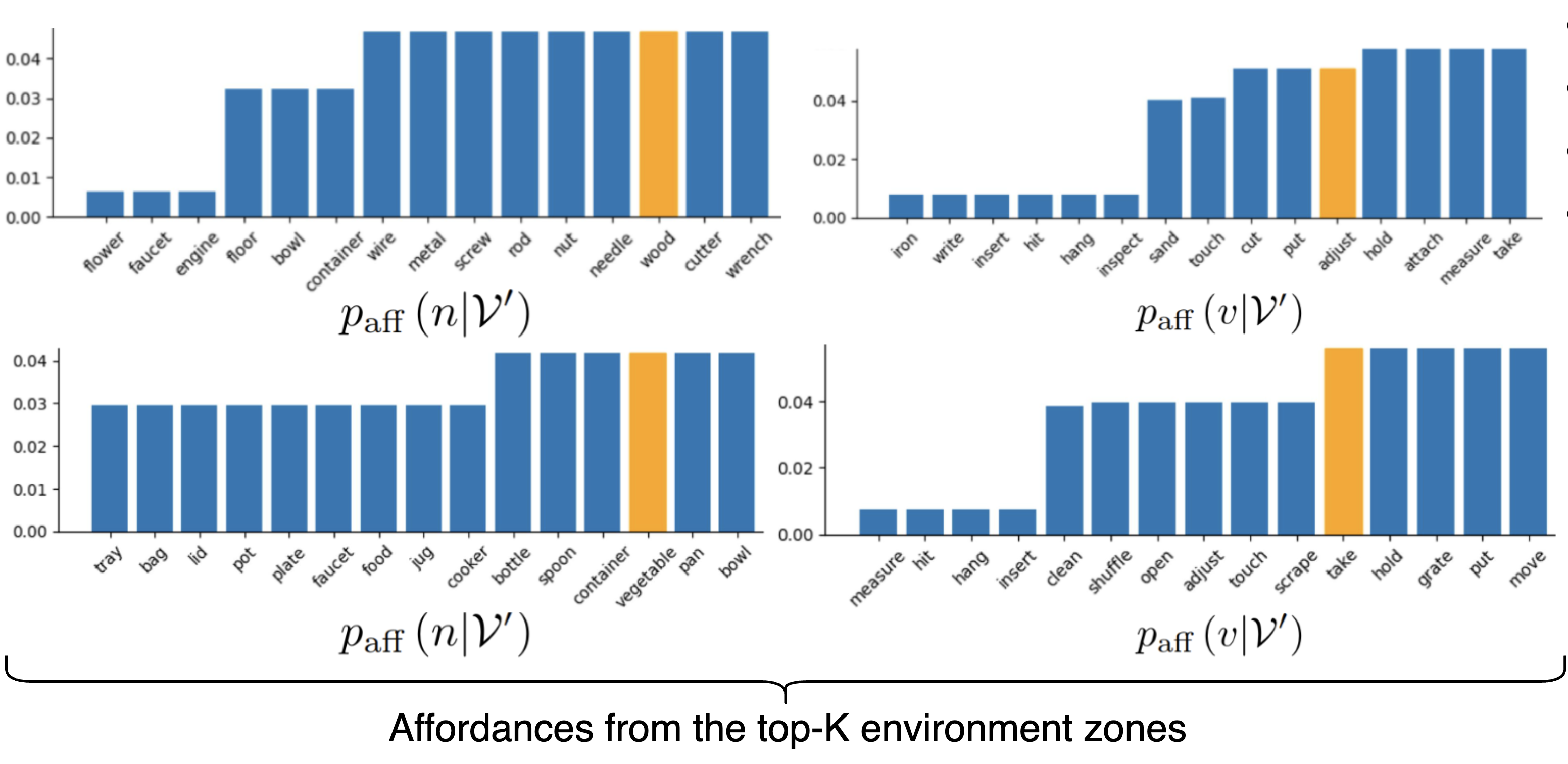}
\includegraphics[width=0.35\textwidth]{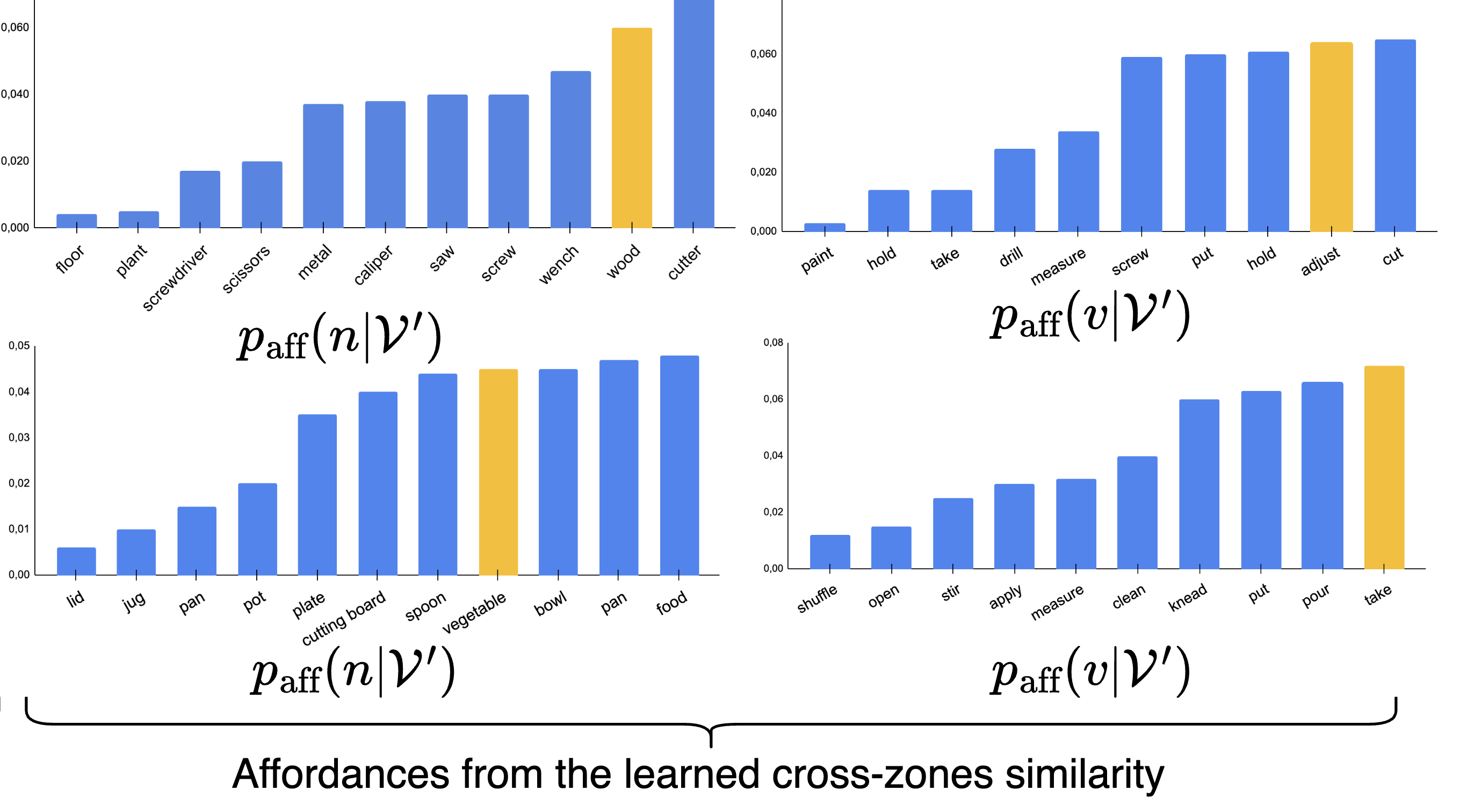}
\caption{\textbf{\textcolor{red}{Comparative between the fixed affordance distribution inferred from top-K environment zones (left) vs. learning the cross-zone similarity (right).}} We also visualize the closest environments in terms of the visual $\mathcal{K}^\mathcal{V}$ and narrative $\mathcal{K}^\mathcal{T}$ cosine similarity. We show in orange the STA ground-truth label, \textcolor{red}{showing that the affordances distribution effectively captures the STA action. Learning the cross-zone similarity enables a more flexible affordance distribution representation, as we do not rely on a fixed number of zones.}}
\label{fig:aff_results}
\end{figure*}

\noindent
\textbf{Temporal Pooling (Table~\ref{tab:staformer_ablation}, Exp C).} 
We start showing the effects of temporal pooling in Experiments B3 (mean temporal pooling), C1 (temporal convolution) and C2 (frame-guided temporal pooling). The temporal convolution helps capturing the video dynamics and obtaining more accurate time to contact estimates, improving N+$\delta$ mAP up to 6.05. However, our frame-guided attention mechanism (Exp C2) enhances spatio-temporal understanding of the video, achieving significant improvements from 8.75 to 10.04 N+V mAP and from 2.94 to 3.39 All mAP. Unlike convolutional pooling, which focuses solely on temporal dynamics, our attention mechanism joins a spatio-temporal understanding of the video by mapping to the 2D reference space of the last observed frame the pooled video features. This advantage extends to DETR head versions, with significant improvements on the time to contact score up to 8.53 N + $\delta$ mAP in Exp C4 (multi-head attention). 
Finally, performing the temporal pooling per-scale further enhances performance, achieving up to 32.58 N mAP, 15.00 N+V mAP, 8.53 N+$\delta$ mAP, and 4.71 All mAP, demonstrating more robust spatio-temporal feature learning adaptable to object sizes.

\noindent
\textbf{Feature Fusion (Table~\ref{tab:staformer_ablation}, Exp D).} 
Experiments D1-D7 of Table~\ref{tab:staformer_ablation} compare the contribution of the proposed Dual Image-Video Attention module for 2D-3D feature fusion.
Comparing experiments D1 vs. C2 shows small but consistent gains when dual image-video attention is used for fusion in STAformer, as compared to simple sum fusion (20.08 vs. 19.78 N, 10.21 vs. 10.04 N + V, 6.51 vs. 6.35 N + $\delta$ and 3.47 vs. 3.39 All mAP).
However, using cross-attention only with image tokens ($I\to\mathcal{V}$-Exp.D3) or video tokens ($\mathcal{V} \to I$ -Exp.D4) as queries, performs worse than the proposed dual image-video attention (Exp. D2), suggesting the need to incorporate the refinement of both modalities.
Incorporating multi-head attention on the temporal pooling and on the 2D-3D fusion (Exp.D5) produces a consistent improvement in all the metrics due to its ability to capture diverse patterns simultaneously. 
However, we do not see any systematic improvement when we apply the MH.Dual Cross Attention on the STAformer++ model. Since we are operating at multi-scale levels, this introduces a long sequence of tokens that makes this mechanism very computational consuming, leading a trade-off between the number of fine-tuned video blocks and the dual cross attention.

\noindent
\textbf{Dependence on video length (Figure \ref{fig:time_temp_pool}).} We analyze the performance with different time windows for the video model, as Figure \ref{fig:time_temp_pool} shows. We compare a temporal pooling with two versions of our frame-guided temporal pooling: the Conv. Temp.Pooling uses convolutional weights on the $Q_{TP}, K_{TP}, V_{TP}$ projection layers, while the Linear Temp.Pooling adopts linear layers plus a positional encoding. 
Averaging features of longer videos reduces the spatial alignment due to the camera movement. However, computing our frame-guided temporal attention pooling projects the video features into the last frame via the attention mechanism, capturing better aligned spatio-temporal features.
This effect is visualized in the N + $\delta$ and Overall mAP plots: while a larger time-window degrades the performance when computing the temporal mean, it benefits the temporal reasoning of the model, obtaining better results when the video covers 1.5 secs rather than 0.5 sec.

\begin{table}[]
    \centering
    \caption{Comparative of the affordances priors effect on Stillfast. Results in mAP on the validation split of Ego4D v1. \textbf{Best results} in bold. Relative gain is respect with the \underline{base model}.}
    \begin{tabular}{|cc|c|c|c|c|}
    \hline
    \multicolumn{1}{|c|}{} &  & N     & N+V   & N + TTC & All \\ \hline
    \multicolumn{1}{|c|}{-}    & -  & \underline{16.20} & \underline{7.47} & \underline{4.94}    & \underline{2.48}    \\ \hline
    \multicolumn{1}{|c|}{\multirow{4}{*}{\begin{tabular}[c]{@{}c@{}}Env.\\ Aff.\end{tabular}}} & Count-based priors & 16.44& 7.84& 4.50& 2.39\\
    \multicolumn{1}{|c|}{}   & \textcolor{red}{Inverse-freq.Prior} & \textcolor{red}{15.22} & \textcolor{red}{6.98}  & \textcolor{red}{4.40}  & \textcolor{red}{2.28}    \\
    \multicolumn{1}{|c|}{}   & Ego-Topo\cite{nagarajan2020ego} & 14.92 & 6.45  & 4.01    & 2.14    \\
    \multicolumn{1}{|c|}{}   & Fixed Weighted (Ours) & 18.44 & 8.46 & 5.47 & 2.85    \\ \hline
    \multicolumn{1}{|c|}{\multirow{3}{*}{\begin{tabular}[c]{@{}c@{}}Int.\\ Hot\end{tabular}}}  & Center Prior        & 14.44& 6.86& 3.90& 2.05\\ 
    \multicolumn{1}{|c|}{}                                                                     & Hands Proximity    & 13.86& 6.15& 3.71& 1.86\\  
    \multicolumn{1}{|c|}{}     & Ours      & 17.82 & 7.62 & 5.05    & 2.53    \\ \hline
    \multicolumn{2}{|c|}{Both (Ours)}                      & \textbf{19.34} & \textbf{8.58} & \textbf{5.55}    & \textbf{2.95}    \\ 
    \multicolumn{2}{|c|}{Gain}     & \textcolor{ForestGreen}{+19.3} & \textcolor{ForestGreen}{+14.9} & \textcolor{ForestGreen}{+12.4}    & \textcolor{ForestGreen}{+18.9}    \\ \hline
    \end{tabular}
    \label{tab:priors}
\end{table}

\begin{table}[t]
\centering  
\caption{Comparative of the affordances priors effect on STAformer. Results in mAP on the validation split of Ego4D v1. \textbf{Best results} in bold. Relative gain is respect with the \underline{base model}.}
\label{tab:staformer_aff}
\begin{tabular}{|cc|c|c|c|c|}
\hline
\multicolumn{1}{|c|}{} &  & N     & N+V   & N + TTC & All \\ \hline
\multicolumn{1}{|c|}{-}    & -  & \underline{21.71} & \underline{10.75} & \underline{7.24}    & \underline{3.53}    \\ \hline
\multicolumn{1}{|c|}{\multirow{4}{*}{\begin{tabular}[c]{@{}c@{}}Env.\\ Aff.\end{tabular}}} & Count-based Prior & 21.96 & 10.98 & 6.80    & 3.56    \\ 
\multicolumn{1}{|c|}{}    & \textcolor{red}{Inverse-freq. Prior} & \textcolor{red}{22.40} & \textcolor{red}{10.27} & \textcolor{red}{6.02}  & \textcolor{red}{2.84}    \\
\multicolumn{1}{|c|}{}                                                                     & Ego-Topo\cite{nagarajan2020ego} & 17.21 & 8.45  & 5.32    & 2.64    \\ 
\multicolumn{1}{|c|}{}                                                                     & Fixed Weighted (Ours) & 23.55 & 11.75 & 7.55    & 3.74    \\ \hline
\multicolumn{1}{|c|}{\multirow{3}{*}{\begin{tabular}[c]{@{}c@{}}Int.\\ Hot\end{tabular}}}  & Center Prior        & 17.70 & 8.82  & 5.22    & 2.62    \\ 
\multicolumn{1}{|c|}{}                                                                     & Hands Proximity    & 16.35 & 7.91  & 4.49    & 2.30    \\ 
\multicolumn{1}{|c|}{}                                                                     & Ours                                                          & 23.63 & 11.38 & 7.51    & 3.66    \\ \hline
\multicolumn{2}{|c|}{Both (Ours)}                                                                                                                          & \textbf{24.36} & \textbf{12.00} & \textbf{7.66}    & \textbf{3.77}    \\ 
\multicolumn{2}{|c|}{Gain}     & \textcolor{ForestGreen}{+12.2} & \textcolor{ForestGreen}{+11.6} & \textcolor{ForestGreen}{+5.8}    & \textcolor{ForestGreen}{+6.8}    \\ \hline
\end{tabular}
\end{table}



\begin{table}[t]
\centering  
\caption{Comparative of the affordances priors effect on STAformer++. Results in mAP on the validation split of Ego4D v1. \textbf{Best results} in bold. Relative gain is respect with the \underline{base model}.}
\label{tab:aff_ablation}
\begin{tabular}{|cc|c|c|c|c|}
\hline
\multicolumn{1}{|c|}{} &  & N     & N+V   & N + TTC & All \\ \hline
\multicolumn{1}{|c|}{-}    & -  & \underline{32.07} & \underline{15.00} & \underline{8.53} & \underline{4.31} \\ \hline
\multicolumn{1}{|c|}{\multirow{5}{*}{\begin{tabular}[c]{@{}c@{}}Env.\\ Aff.\end{tabular}}} & \textcolor{red}{Count-based Prior} & \textcolor{red}{31.39} & \textcolor{red}{13.68} & \textcolor{red}{8.71} & \textcolor{red}{4.42} \\
\multicolumn{1}{|c|}{}    & \textcolor{red}{Inverse-freq. Prior} & \textcolor{red}{31.67} & \textcolor{red}{13.42} & \textcolor{red}{8.48} & \textcolor{red}{4.17} \\
\multicolumn{1}{|c|}{}   & \textcolor{red}{Ego-Topo\cite{nagarajan2020ego}} & \textcolor{red}{28.39} & \textcolor{red}{12.02}  & \textcolor{red}{7.72} & \textcolor{red}{3.78}    \\ 
\multicolumn{1}{|c|}{}   & Fixed Weighted & 31.00 & 14.17 & 8.69 & 3.84 \\ 
\multicolumn{1}{|c|}{}   & Learned & \textbf{33.21} & \textbf{15.94} & \textbf{8.98} & \textbf{4.66} \\ \hline
\multicolumn{1}{|c|}{\multirow{3}{*}{\begin{tabular}[c]{@{}c@{}}Int.\\ Hot\end{tabular}}}  & \textcolor{red}{Center Prior} & \textcolor{red}{28.12} & \textcolor{red}{13.77}  & \textcolor{red}{7.89}  & \textcolor{red}{3.89}    \\ 
\multicolumn{1}{|c|}{}   & \textcolor{red}{Hands Proximity}    & \textcolor{red}{27.55} & \textcolor{red}{13.29}  & \textcolor{red}{7.12}  & \textcolor{red}{3.54}    \\

\multicolumn{1}{|c|}{} & \textcolor{red}{Ours} & \textcolor{red}{32.84} & \textcolor{red}{15.67}  & \textcolor{red}{8.67}  & \textcolor{red}{4.35}    \\ \hline
\multicolumn{2}{|c|}{\textcolor{red}{Both (Ours)}}  & \textbf{\textcolor{red}{33.44}} & \textbf{\textcolor{red}{16.15}} & \textbf{\textcolor{red}{9.06}}    & \textbf{\textcolor{red}{4.78}}    \\ 

\multicolumn{2}{|c|}{Gain}  & \textcolor{ForestGreen}{+4.2} & \textcolor{ForestGreen}{+7.6} & \textcolor{ForestGreen}{+6.2}    & \textcolor{ForestGreen}{+10.9}    \\ \hline
\end{tabular}
\end{table}

\begin{figure}[t]
    \centering
    \includegraphics[width=0.31\columnwidth]{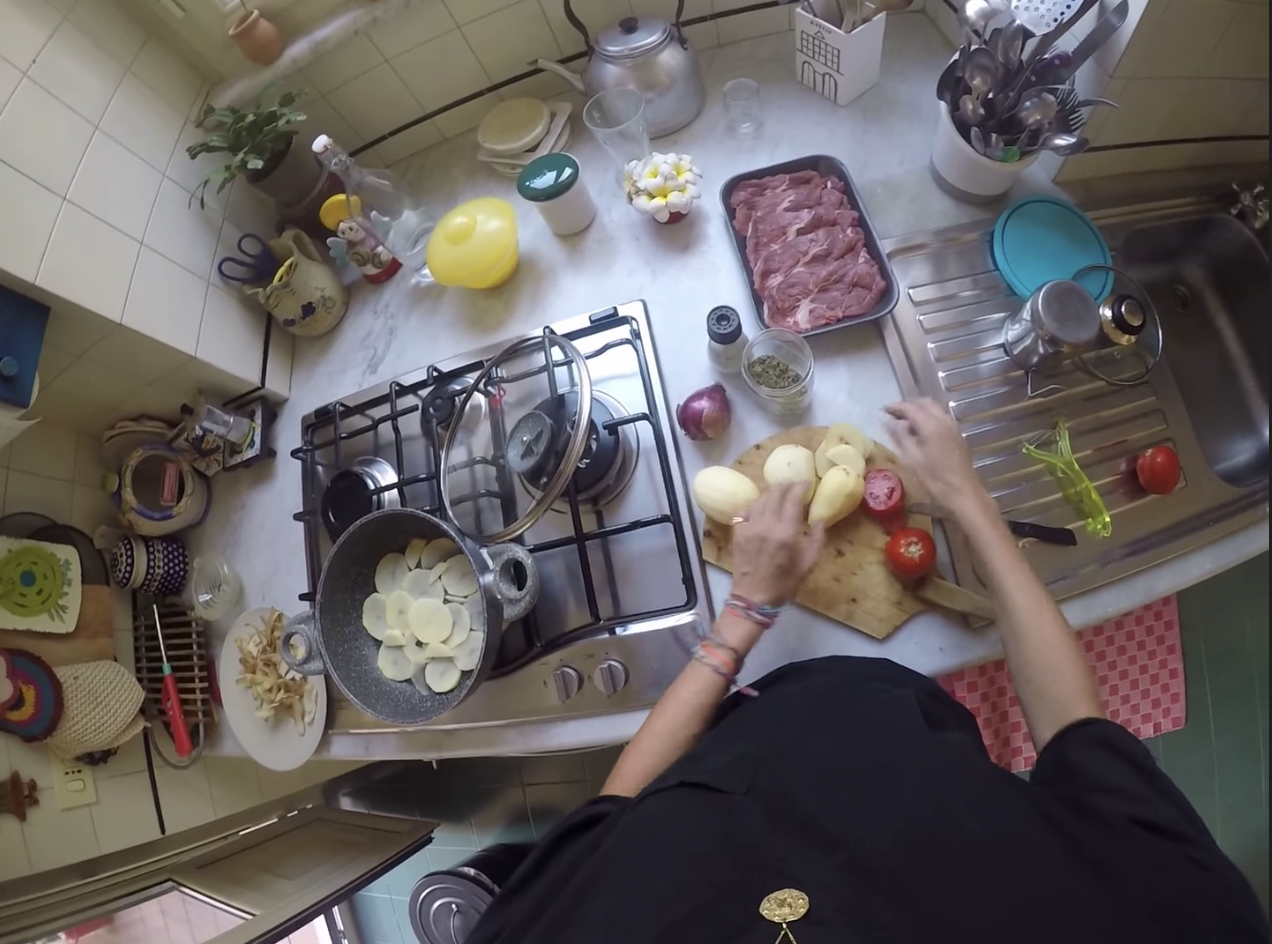}
    \includegraphics[width=0.31\columnwidth]{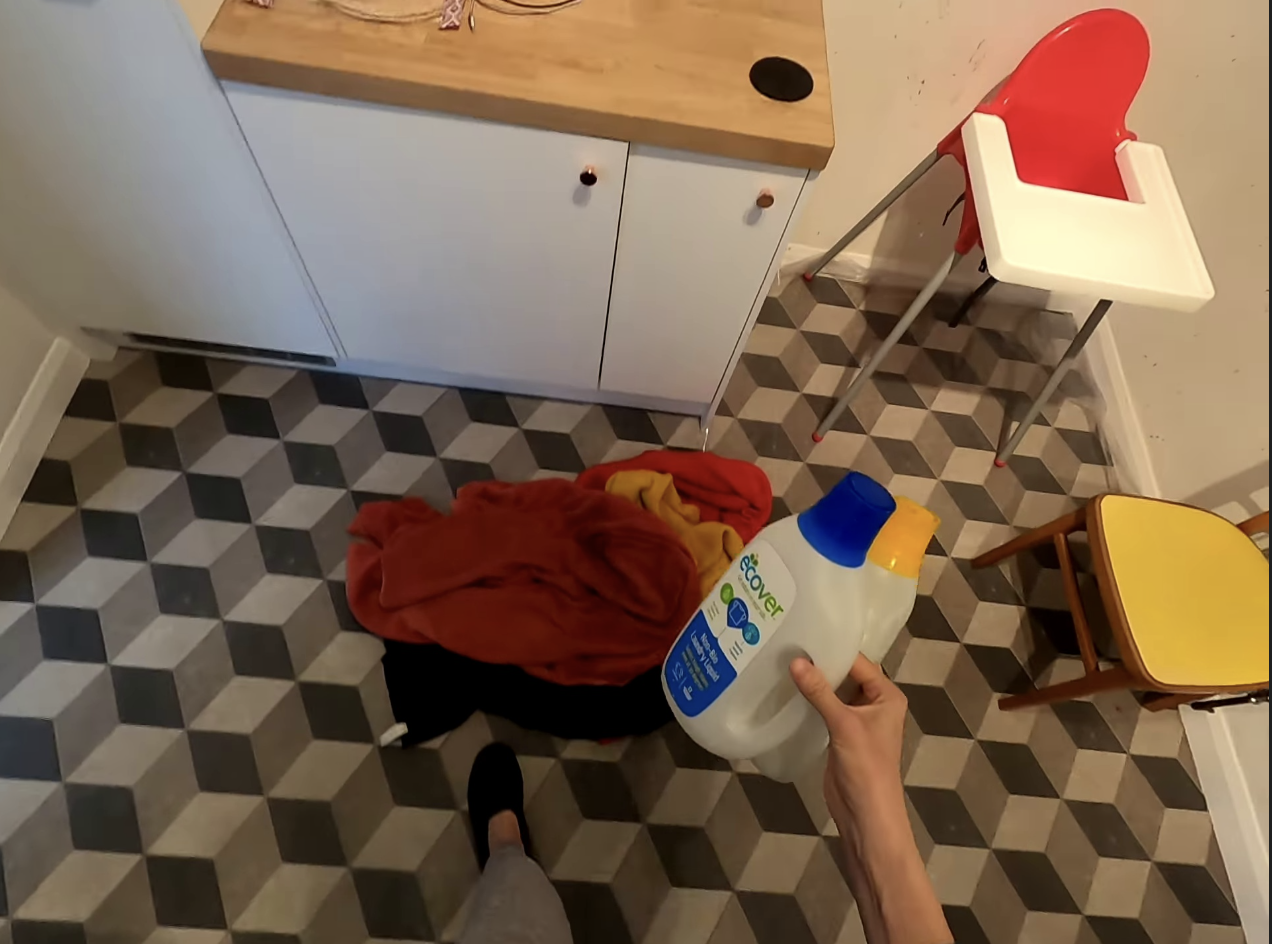}
    \includegraphics[width=0.31\columnwidth]{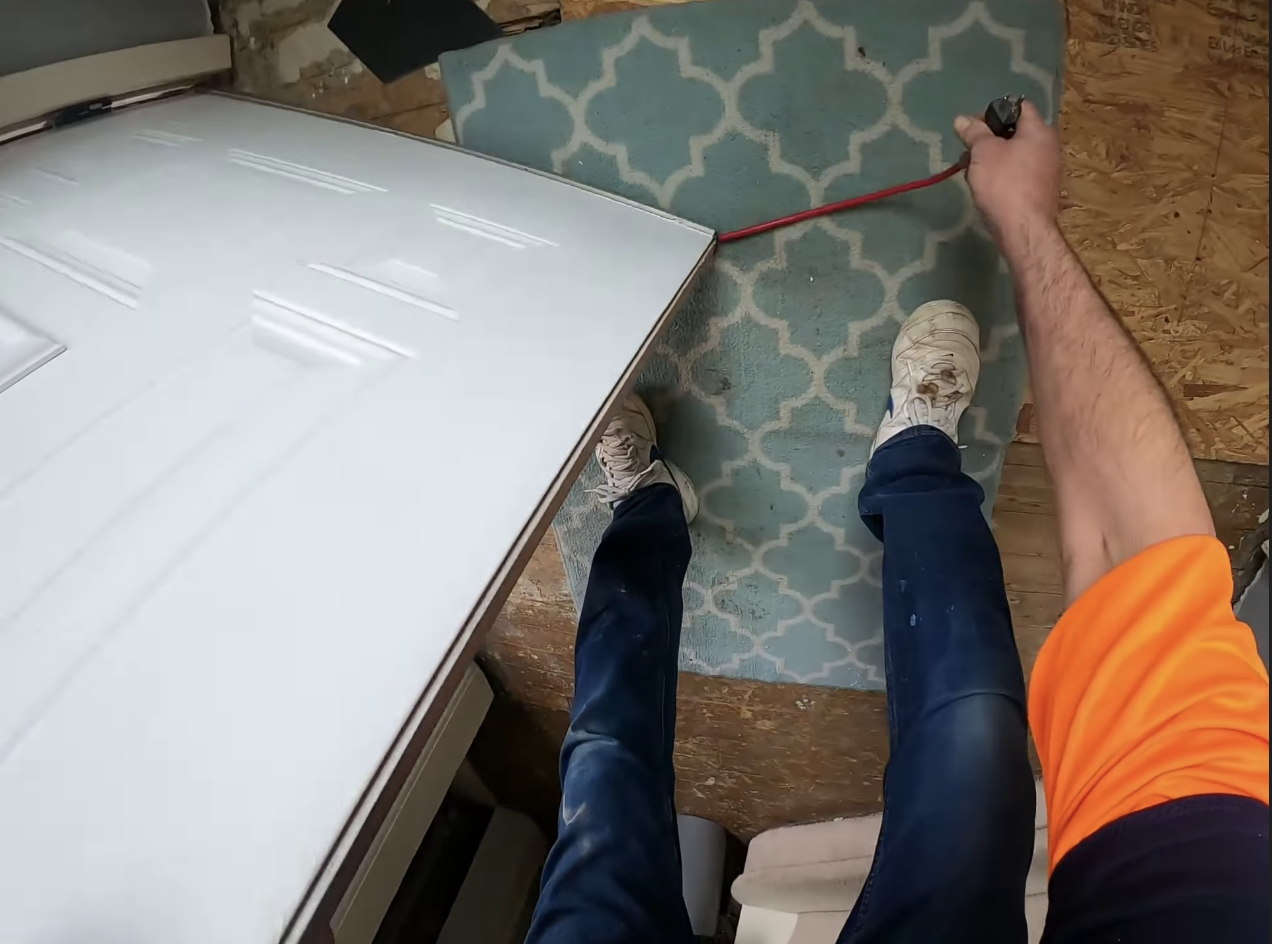}
    \\
    \centering
    \includegraphics[width=0.31\columnwidth]{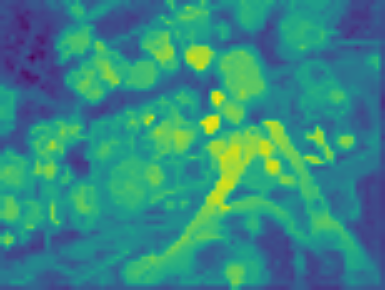}
    \includegraphics[width=0.31\columnwidth]{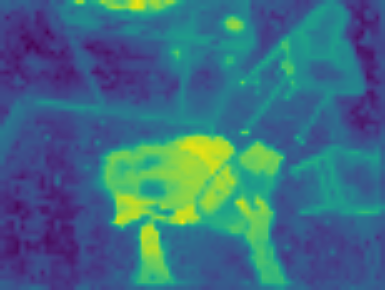}
    \includegraphics[width=0.31\columnwidth]{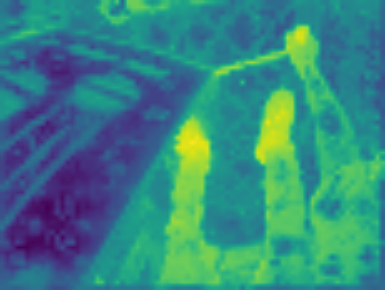}
    \\
    \centering
    \includegraphics[width=0.31\columnwidth]{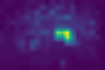}
    \includegraphics[width=0.31\columnwidth]{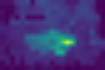}
    \includegraphics[width=0.31\columnwidth]{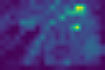}
    \\
    \centering
    \includegraphics[width=0.31\columnwidth]{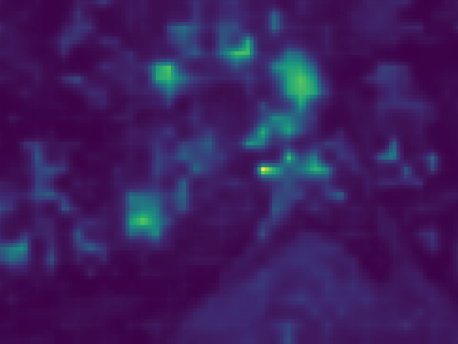}
    \includegraphics[width=0.31\columnwidth]{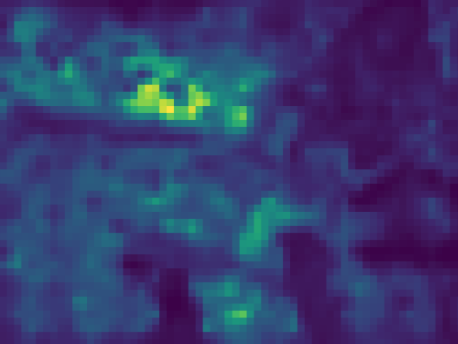}
    \includegraphics[width=0.31\columnwidth]{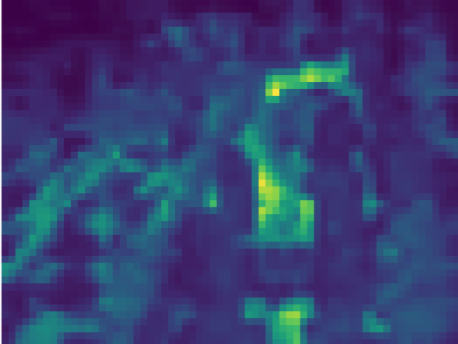}
    \\
    \centering
    \includegraphics[width=0.25\columnwidth]{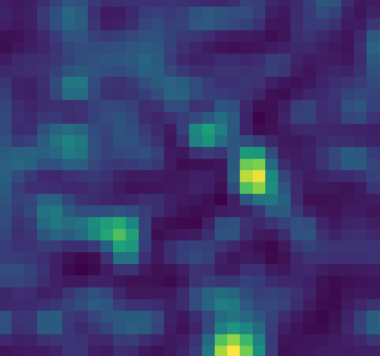} \hspace{0.45cm}
    \includegraphics[width=0.25\columnwidth]{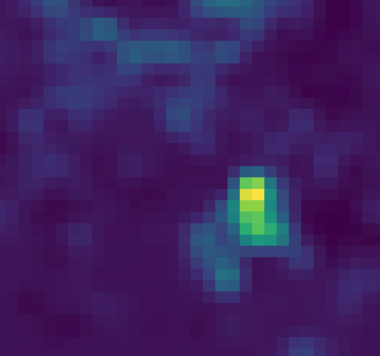} \hspace{0.45cm}
    \includegraphics[width=0.25\columnwidth]{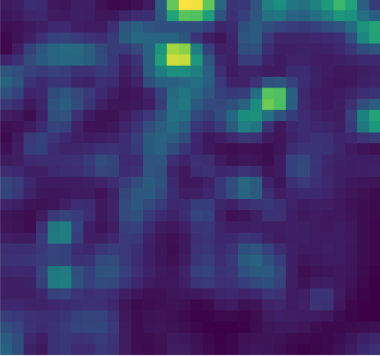}
    \\

    \caption{\textbf{Dual image-video attention maps, qualitative results.}
    Top to bottom: STAformer final predictions, attention map of pooled TimeSformer tokens (queries) on DINOv2 image tokens (keys and values), attention of DINOv2 image tokens (queries) on pooled TimeSformer  video tokens (keys and values), attention map of pooled EgoVideo tokens (queries) on intermediate Swin-T image tokens (keys and values) and attention of intermediate Swim-T image tokens (queries) on pooled EgoVideo features. 
    Video tokens attend fine-grained object information from the high-resolution image; image features focus on objects which are important for future interactions.}
    \label{fig:dual_cross_img}
\end{figure}

\begin{figure*}[t]
    \centering
    \includegraphics[width=0.195\textwidth]{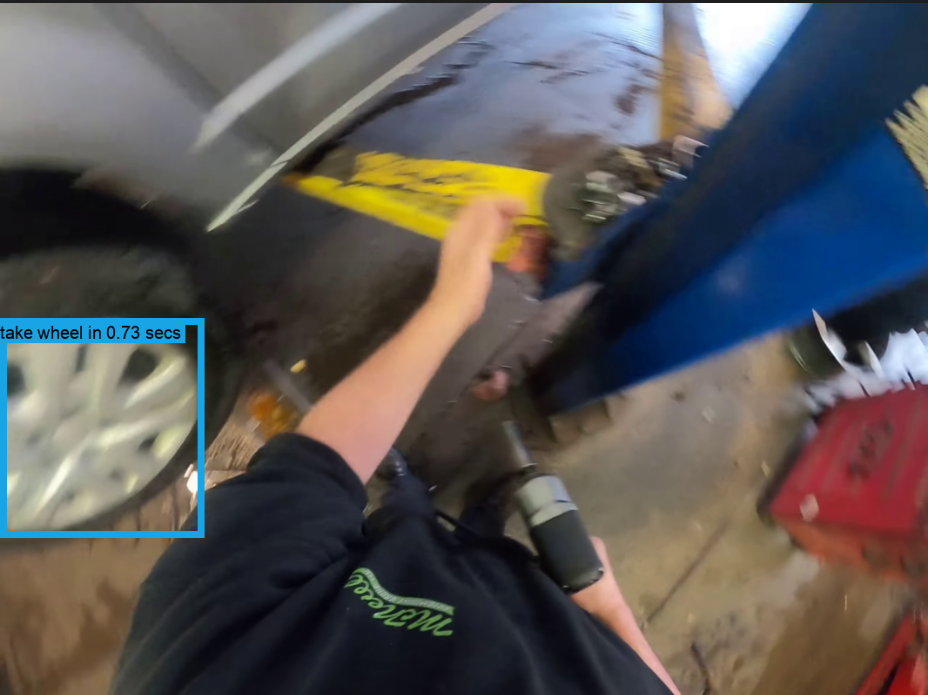}
    \includegraphics[width=0.195\textwidth]{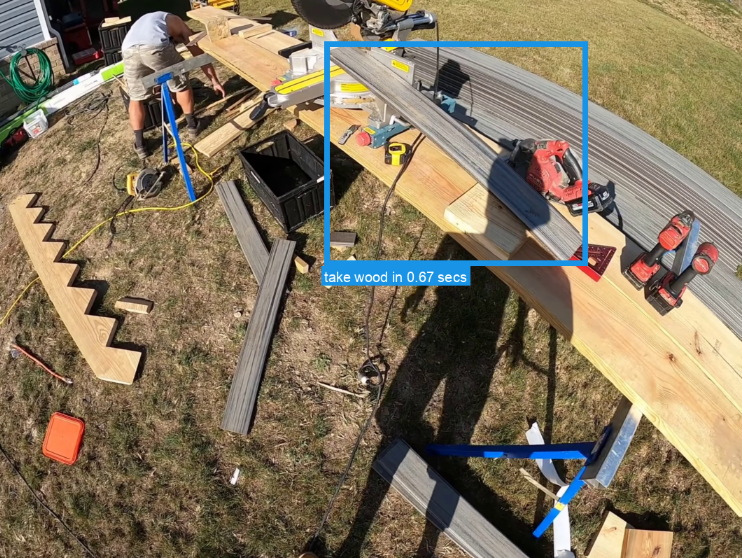}
    \includegraphics[width=0.195\textwidth]{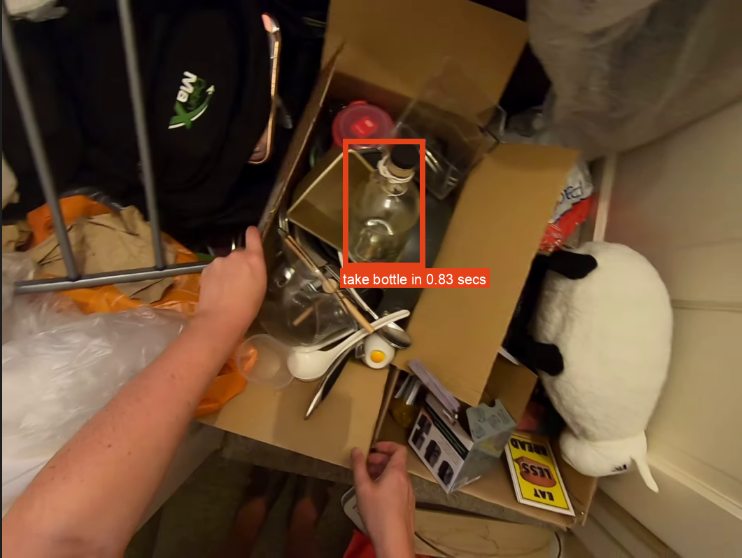}
    \includegraphics[width=0.195\textwidth]{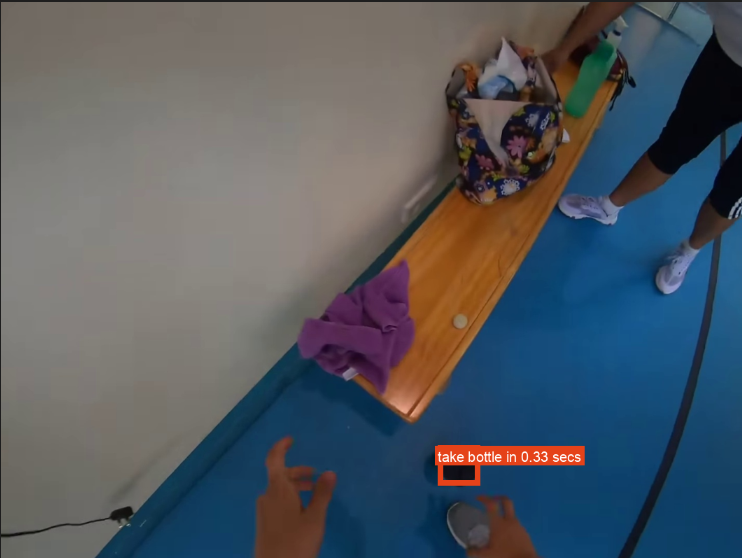}
    \includegraphics[width=0.195\textwidth]{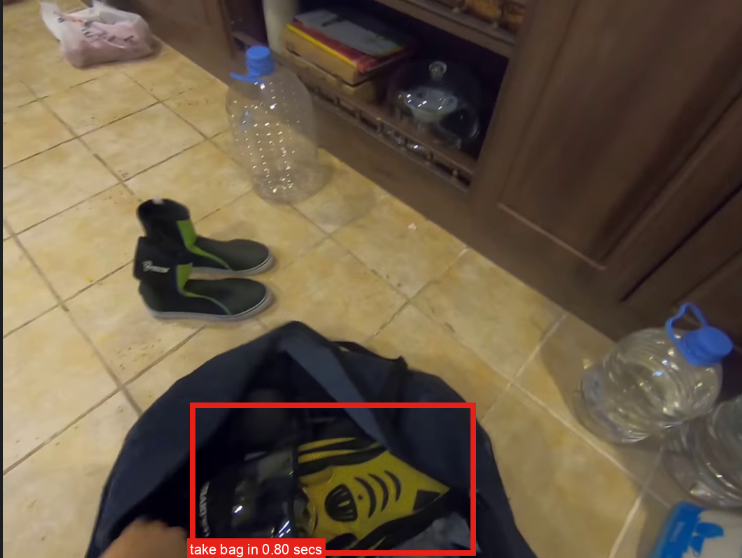}
    \\
    \centering
    \includegraphics[width=0.195\textwidth]{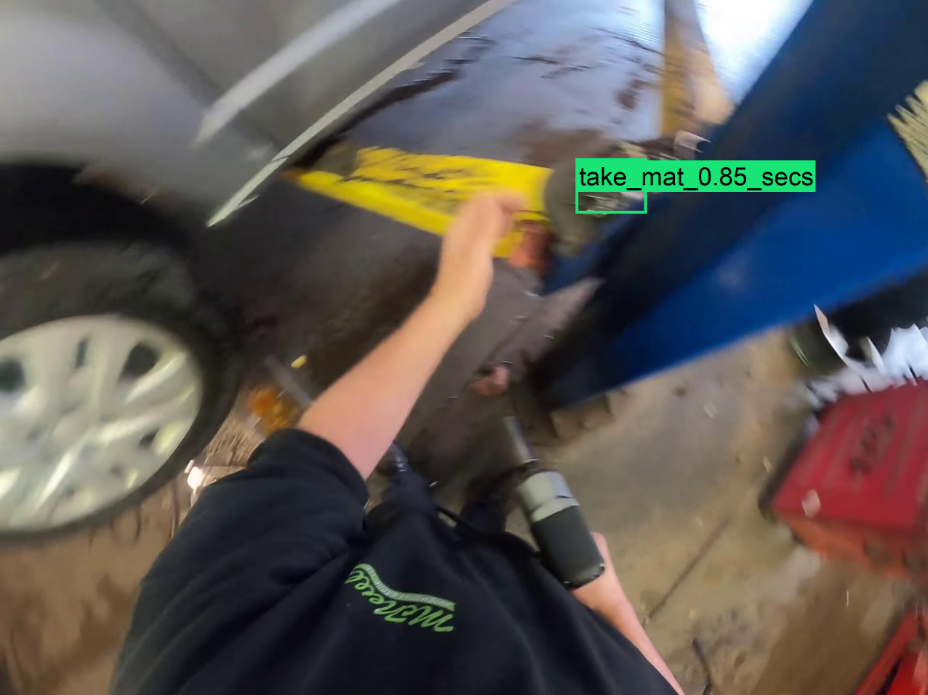}
    \includegraphics[width=0.195\textwidth]{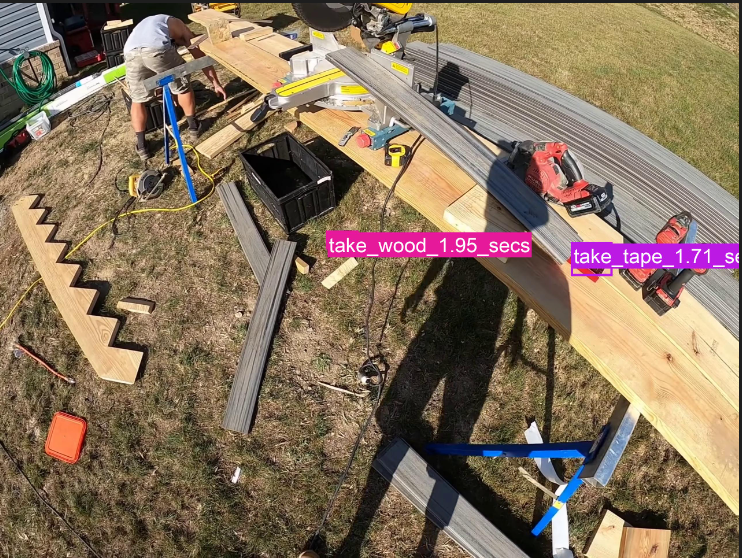}
    \includegraphics[width=0.195\textwidth]{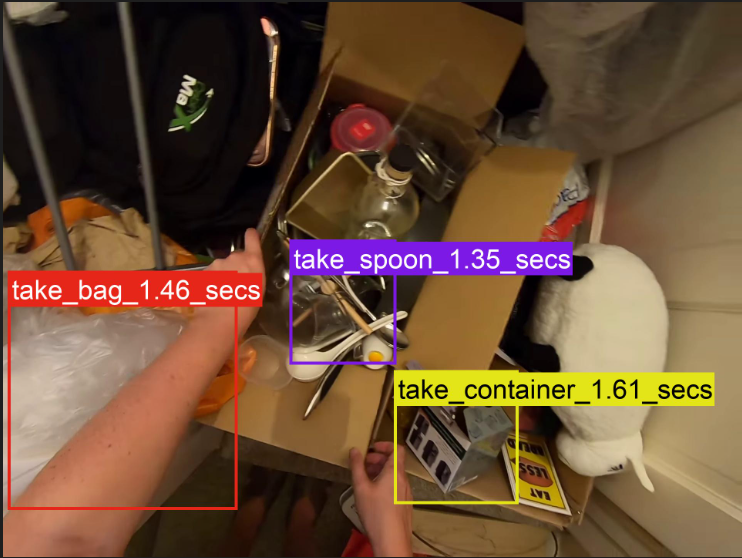}
    \includegraphics[width=0.195\textwidth]{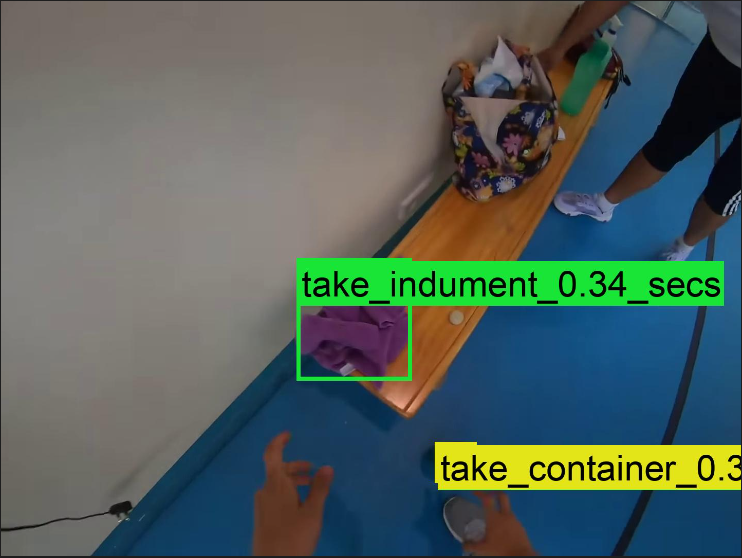}
    \includegraphics[width=0.195\textwidth]{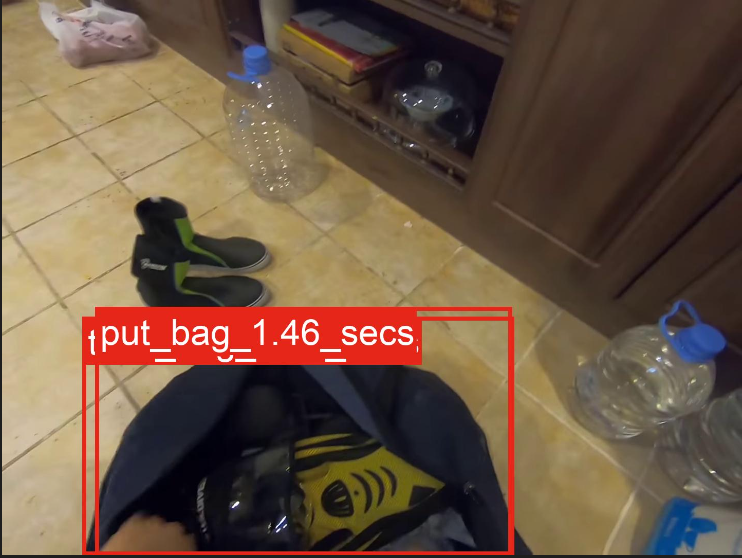}
    \\
    \centering
    \includegraphics[width=0.195\textwidth]{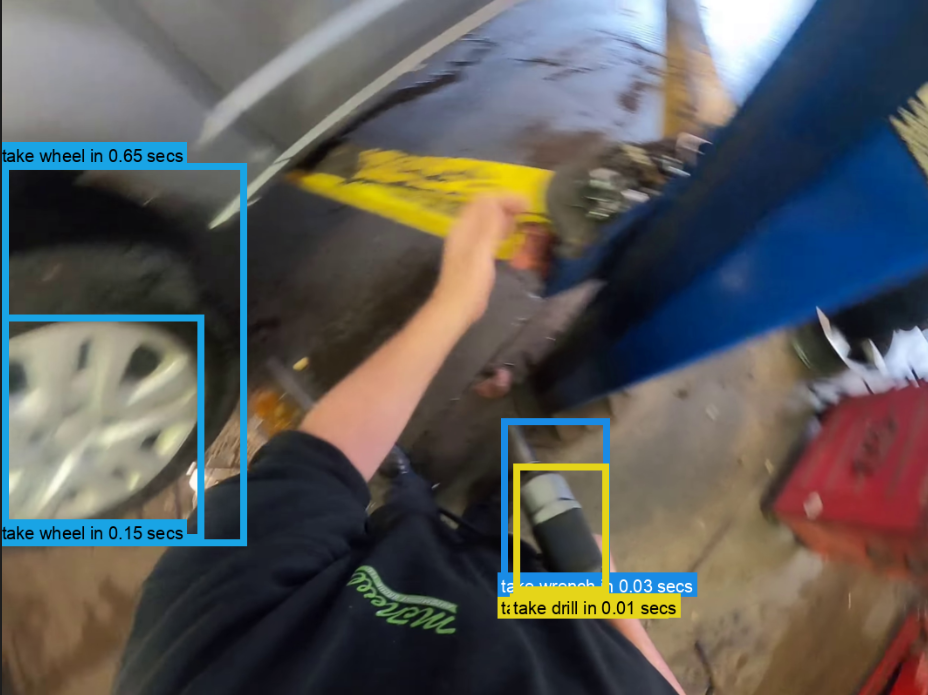}
    \includegraphics[width=0.195\textwidth]{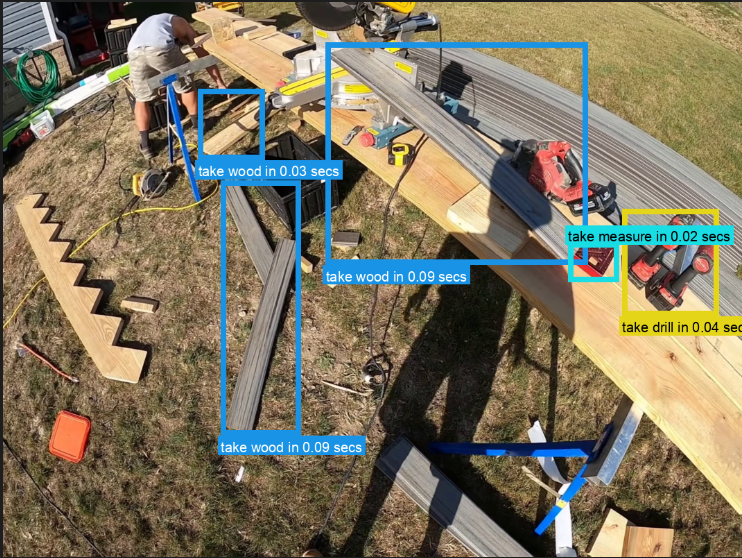}
    \includegraphics[width=0.195\textwidth]{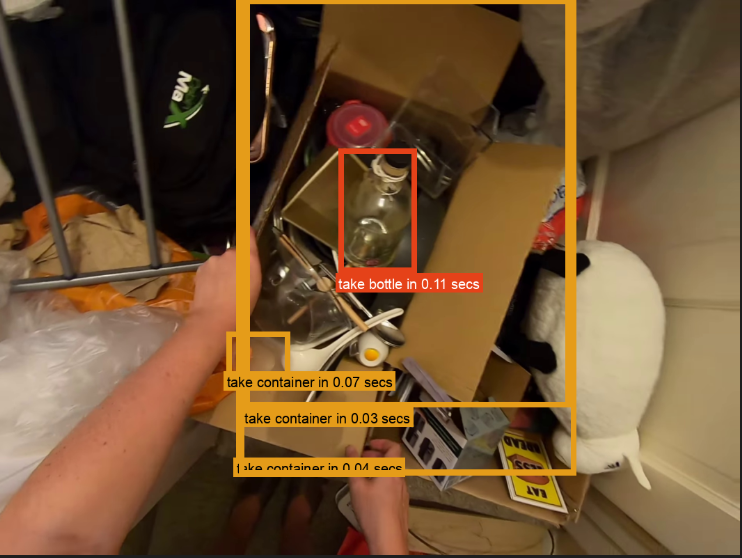}
    \includegraphics[width=0.195\textwidth]{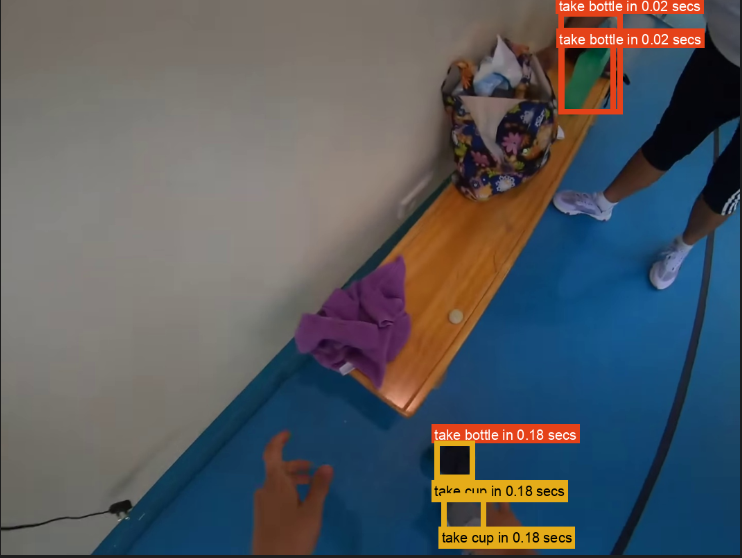}
    \includegraphics[width=0.195\textwidth]{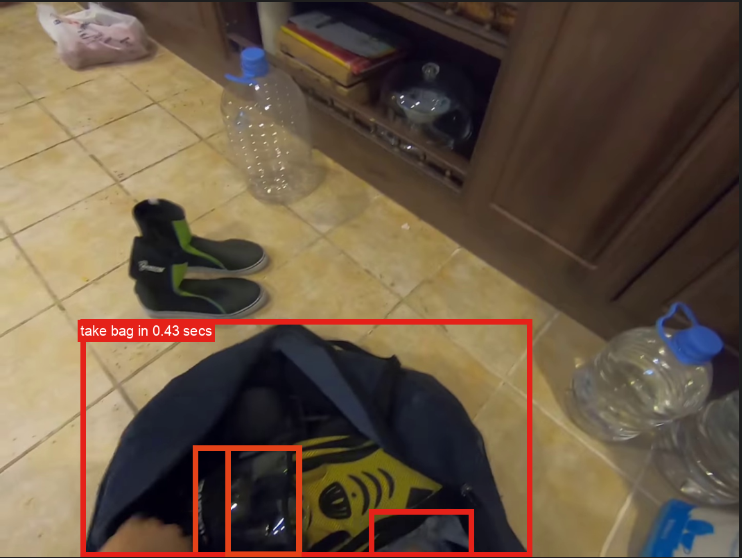}

    \caption{\textbf{Ego4D Qualitative results}
    Top to bottom: ground truth, STAformer predictions and STAformer++ predictions in Ego4D v2 validation split. We visualize the top-5 detections by the model. It is appreciated how the STAformer++ detections capture better the contour of the object, and that the whole model achieves a better understanding of the potential interactions in the video.}
    \label{fig:Ego4D_qualit}
\end{figure*}

\subsection{Ablation Study on Affordances}

Tables \ref{tab:priors}, \ref{tab:staformer_aff}, \ref{tab:aff_ablation} detail the influence of Environment Affordances (E.AFF) and Interaction Hotspots (I.H), when integrated, separately and jointly, showing in all cases consistent improvements. 

\noindent
\textbf{Environment Affordances on Stillfast (Table \ref{tab:priors}) and STAformer  (Table \ref{tab:staformer_aff}).}
We first evaluate a naive Count-based Prior, re-weighting nouns and verbs probabilities by their frequency in the training dataset. While it slightly improves some metrics, it highlights the need to relate test samples to the specific scene's affordances. 
\textcolor{red}{We also evaluate an inverse-frequency prior, which enhances performance for tail noun and verb classes, although overall results remain suboptimal.}
Training a NN classifier as in \cite{nagarajan2020ego} does not produce a useful distribution of the affordances for fusion with STA probabilities. Our intuition is that the NN overfits to the interactions in the scene which are more obvious, losing the generalist quality of our predictions across environments. 
Our Fixed Env.Aff approach significantly refines nouns and verbs probabilities, obtaining consistent gains in $N+V$ Top-5 mAP (8.45 vs. 7.47 in Stillfast and 11.75 vs. 10.75 in STAformer). 
Figure \ref{fig:aff_results} \textcolor{red}{shows the noun and verb affordance distributions obtained from our two proposed methods.}
Although the ground truth STA class is not top-ranked, it appears in both predicted verb and noun affordances, supporting the observation that similar scenes afford similar interactions.
\textcolor{red}{Comparing both distributions, learning a cross-zone similarity yields a more adaptive and flexible affordance representation, as each affordance class is computed independently from the others through a max-similarity operation across the regions containing the interaction.}

\noindent
\textbf{Interaction Hotspots on Stillfast (Table \ref{tab:priors}),  STAformer (Table \ref{tab:staformer_aff}) \textcolor{red}{and STAformer ++ (Table \ref{tab:aff_ablation})}.}
We start evaluating the interaction hotspots with simple spatial priors. 
A center prior, that benefits bounding box predictions in the center of the scene, is detrimental to performance due to the complexity of egocentric video in which the objects appearing in the peripheral areas can be interacted with in the future.
Similarly, re-weighting based on the current hands location with respect to the object proves ineffective, highlighting the importance of explicitly modeling future hand motion to predict the next interacted objects.
Re-weighing confidence scores based on the spatial prior provided by the interaction hotspots produces a general improvement in all the metrics (e.g., N mAP of 17.82 vs 16.20 in StillFast and 23.63 vs 21.71 in STAformer - mAP All of 2.53 vs 2.48 in StillFast and 3.66 vs 2.53 in STAformer) by accounting for future interaction locations.
\textcolor{red}{The integration of interaction hotspots also enhances STAformer++, increasing the final performance from 32.07 to 32.84 N mAP and from 15.00 to 15.67 N+V mAP.}
Combining environment affordances and hotspots brings significant improvements in both StillFast and STAformer. For instance, STAformer improves N mAP from 21.72 to 24.36 and All mAP from 3.53 to 3.77.

\noindent
\textbf{Learned vs. Fixed Environment Affordances in STAformer++.} 
We compare our approaches for grounding environment affordances in the STA task in Table \ref{tab:aff_ablation}.
Learning affordances during training shows consistent gains in all the metrics, from 32.07 to 33.21 N mAP, 15.00 to 15.94 N+V mAP, 8.53 to 8.98 N+$\delta$ mAP and 4.31 to 4.66 All mAP. 
At higher performance levels, the use of a fixed distribution results in degradation, demonstrating the significance of a flexible and adaptive affordance representation for refining the probabilities.

\subsection{Qualitative results}
Figure \ref{fig:dual_cross_img} reports attention maps produced within the dual image-video attention module and final predictions (top). Video tokens attend fine-grained object information in the high-resolution image (middle), while image tokens attend scene dynamics in video features, which correspond to regions important for future interactions, such as moving hands or objects (bottom).
We illustrate in Figure \ref{fig:Ego4D_qualit} a qualitative comparative on the Ego4D dataset between our two proposed models: STAformer and STAFormer++. The results show qualitatively the improvements achieved our novel architecture version. First, the detected bounding boxes delimit significantly better the objects contour (i.e, the ``wood'' in the second column or the ``bag'' in the final example). Next, they predict more correctly the semantic class of the detected objects (i.e, ``tape'' in the second, or ``container'' in the fourth column). Finally, STAformer++ captures better the action dynamics and offers more plausible next-interactions according to the scene context, as the two first examples show.
\section{Conclusions}
In this paper, we addressed the problem of Short-Term object-interaction Anticipation (STA). 
We proposed two novel architectures for STA. First, STAformer leveraged transformer models for feature extraction, and introduces novel components for image-video fusion, as the frame-guided temporal pooling or the dual cross-attention. 
Next, with STAformer++ we further improve performance by adopting a DETR prediction head.
Our work also explores the contributions of environment affordances and interaction hotspots for refining the probabilities of STA models. We first propose a fixed representation which we exploit at inference with late fusion. A second approach enables STAformer++ to learn to extract the similarity of the current video with a memory of the past interactions in order to extrapolate the affordances distribution.
Our results showcase the improvements given by the proposed architecture and affordance modules, which scores first on all splits of the challenging Ego4D and EPIC-Kitchens benchmarks.
We also detailed the contribution of each individual component through ablations and showed that the integration of affordances is beneficial also to other STA architecture besides the proposed one.
Code and all the material to replicate the results are publicly released to support further research in this area.

\section*{Acknowledgments}

Research at University of Catania has been supported by the project Future Artificial Intelligence Research (FAIR) – PNRR MUR Cod. PE0000013 - CUP: E63C22001940006. Research at the University of Zaragoza was supported by projects PID2021-125209OB-I00 and PID2024-158322OB-I00
(MCIN/ AEI/10.13039/ 501100011033, FEDER/UE and NextGenerationEU/PRTR).

\ifCLASSOPTIONcaptionsoff
  \newpage
\fi

\bibliographystyle{IEEEtran} 
\bibliography{main}

\begin{IEEEbiography}
[{\includegraphics[width=1in,height=1.25in,clip, keepaspectratio]{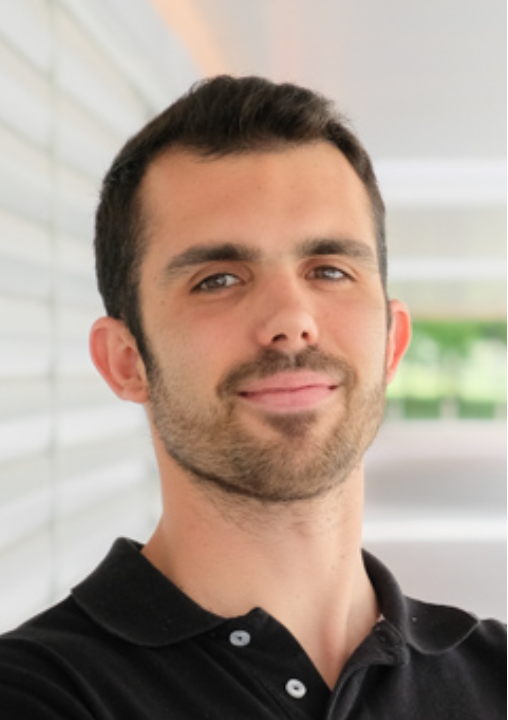}}]{Lorenzo Mur-Labadia}
is a Ph.D. candidate at the University of Zaragoza where he obtained his M.Sc in Robotics, Graphics and Computer Vision in 2022.
He was a visiting researcher at the University of Freiburg (Germany) in 2021 and at the University of Catania (Italy) in 2023.
His research focuses on machine learning and computer vision, particularly scene understanding, object detection, and video analysis, with an interest in multi-modal integration, including language and 3D data.

\end{IEEEbiography}

\begin{IEEEbiography}
[{\includegraphics[width=1in,height=1.25in,clip,keepaspectratio]{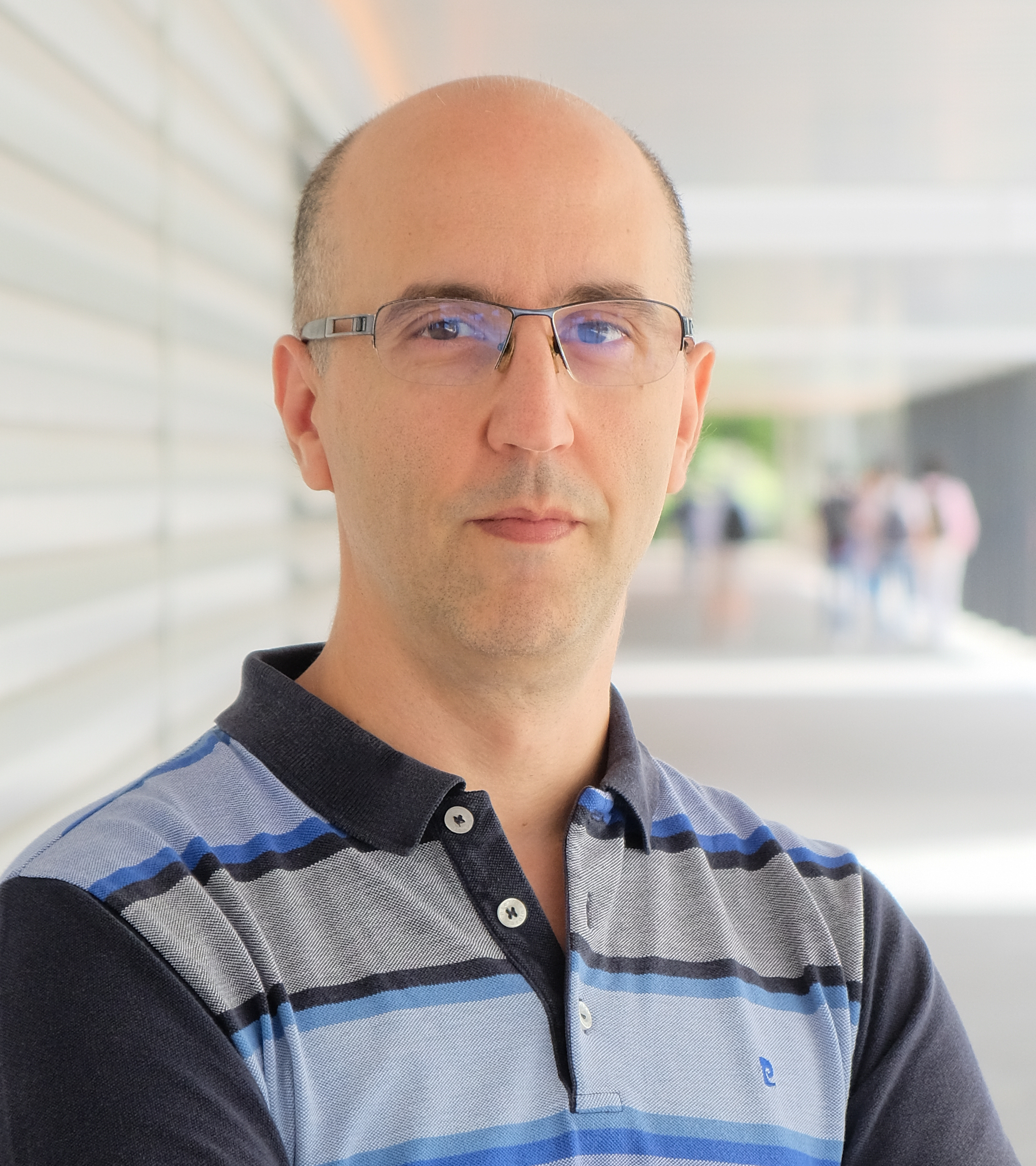}}]
{Ruben Martinez-Cantin}
received his Ph.D. degree from the University of Zaragoza, in 2008. He is currently an Associate Professor at the Department of Computer Science and Systems Engineering, University of Zaragoza. He is also a member of the Robotics, Computer Vision and Artificial Intelligence Group, and the Aragon Institute of Engineering Research (I3A). His current research interests include machine learning, Bayesian inference, computer vision and robotics, particularly in Bayesian optimization and Bayesian deep learning for medical imaging and intelligent assistive devices.
\end{IEEEbiography}

\begin{IEEEbiography} 
[{\includegraphics[width=1in,height=1.25in,clip,keepaspectratio]{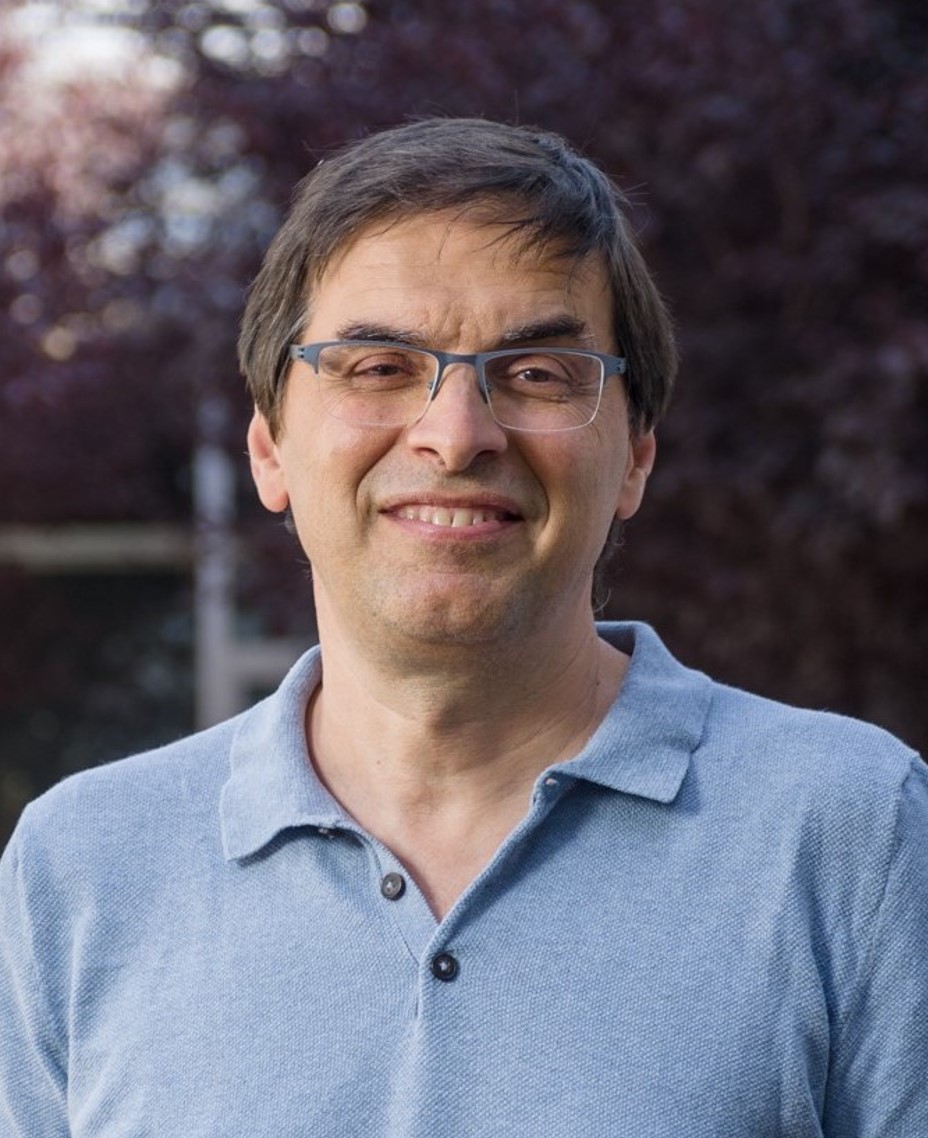}}]
{Josechu Guerrero}
obtained the Ph.D. degree from the University of Zaragoza in 1996. He
is currently Full Professor with the Department
of Computer Science and Systems Engineering,
University of Zaragoza. He is a member of the
Robotics, Computer Vision and Artificial Intelligence
Group, and the Aragon Institute of Engineering
Research (I3A). His current research interests
are in the area of computer vision, particularly
in 3D visual perception, robotics, omnidirectional
vision, vision-based navigation, and the application of computer vision and
robotics techniques to assistive devices.
\end{IEEEbiography}

\begin{IEEEbiography}[{\includegraphics[width=1in,height=1.25in,clip,keepaspectratio]{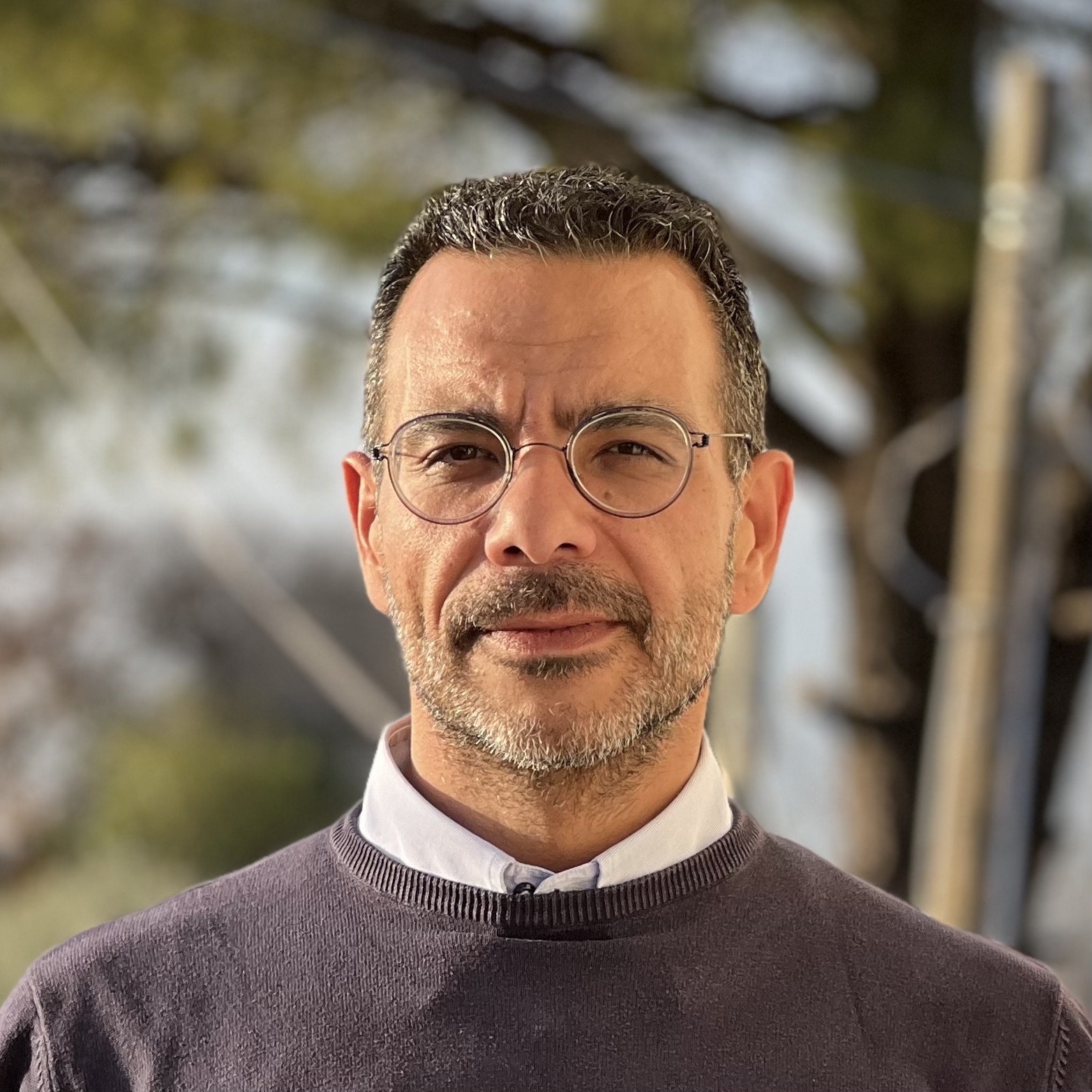}}]{Giovanni Maria-Farinella} (Senior Member, IEEE) is a Full Professor, at the University of Catania, Italy. His research interests lie in the fields of Computer Vision and Machine Learning with focus on Egocentric Vision. Prof. Farinella is part of the EPIC-KITCHENS and EGO4D team. He is Associate Editor of the international journals IEEE Transactions on Pattern Analysis and Machine Intelligence, Pattern Recognition, International Journal of Computer Vision. He has served as Area Chair for CVPR, ICCV and ECCV. He has been Program Chair of ECCV 2022. He founded and currently directs the International Computer Vision Summer School (ICVSS). He was awarded the PAMI Mark Everingham Prize in 2017 and the Intel’s 2022 Outstanding Researcher Award.
\end{IEEEbiography}

\begin{IEEEbiography}[{\includegraphics[width=1in,height=1.25in,clip,keepaspectratio]{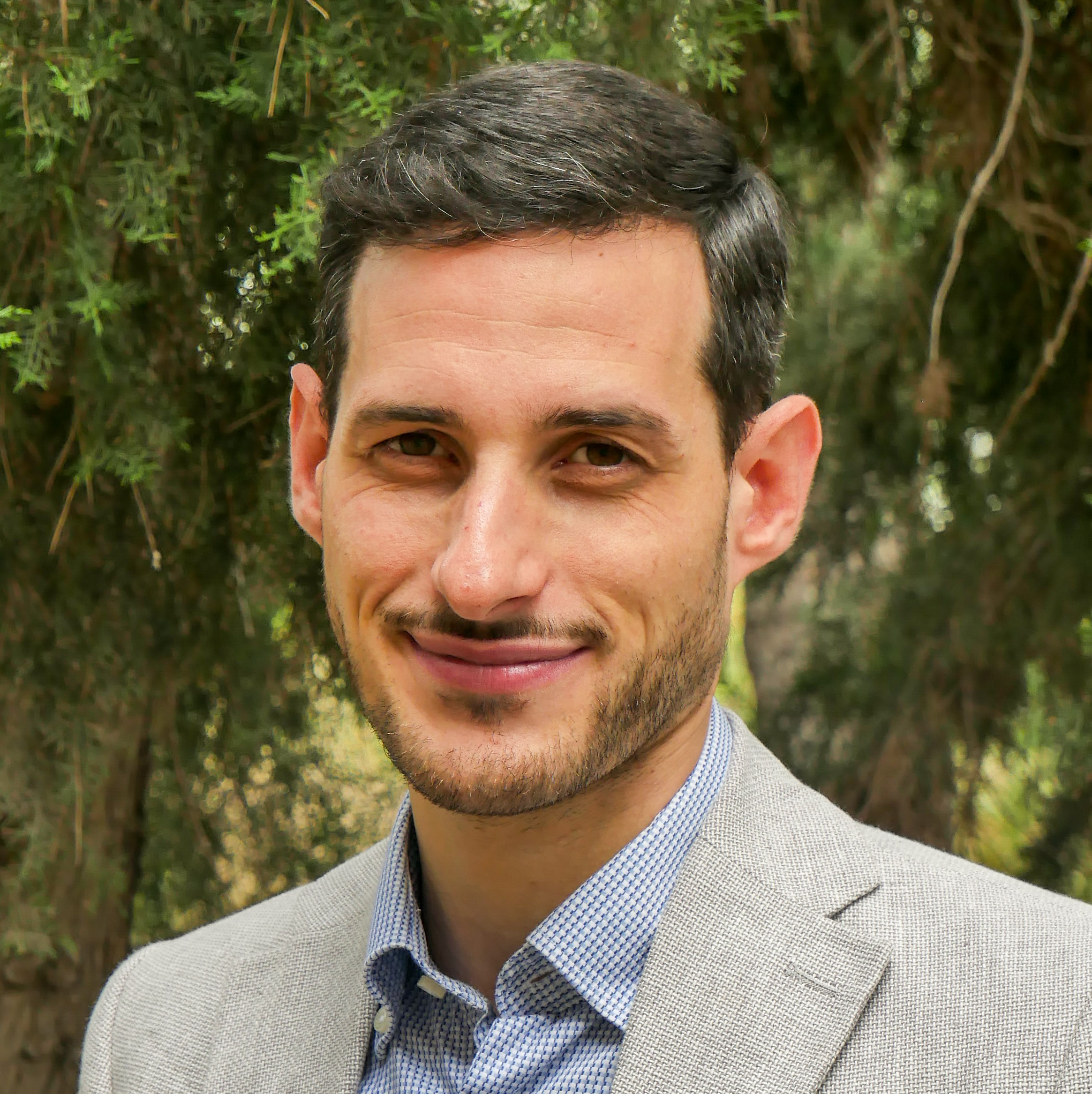}}]{Antonino Furnari} (Senior Member, IEEE) is an Assistant Professor at the University of Catania. He received his Ph.D. in Mathematics and Computer Science from the University of Catania, Italy, in 2017. His research focuses on computer vision, pattern recognition, and machine learning, with a particular emphasis on egocentric (first-person) vision, including scene understanding, object interaction, image-based localization, and visual navigation.
\end{IEEEbiography}





\end{document}